\documentclass[a4paper,fleqn]{cas-dc}

\usepackage[authoryear,longnamesfirst]{natbib}
\usepackage{amsmath,amssymb}
\usepackage{multirow}
\usepackage{tabularx}
\usepackage{flushend}
\usepackage{lipsum}
\usepackage{listings}
\usepackage{caption}
\usepackage[switch]{lineno}
\newtheorem{definition}{Definition}
\def\tsc#1{\csdef{#1}{\textsc{\lowercase{#1}}\xspace}}
\tsc{WGM}
\tsc{QE}
\tsc{EP}
\tsc{PMS}
\tsc{BEC}
\tsc{DE}

\usepackage{balance}

\DeclareCaptionFont{white}{\color{white}}
\DeclareCaptionFormat{listing}{\colorbox[cmyk]{0.43, 0.35, 0.35, 0.01}{\parbox{0.475\textwidth}{\hspace{3pt}#1#2#3}}}
\begin{document}

\let\WriteBookmarks\relax
\def\floatpagepagefraction{1}
\def\textpagefraction{.001}

\shorttitle{Knowledge Representation in Digital Agriculture: A Step Towards Standardised Model}

\shortauthors{Quoc Hung Ngo et~al.}

\title [mode = title]{Knowledge Representation in Digital Agriculture: A Step Towards Standardised Model}                      

\author[1]{Quoc Hung Ngo}[orcid=0000-0001-8246-8392]
\cormark[1]
\ead{hung.ngo@ucdconnect.ie}


\affiliation[1]{organization={University College Dublin},
                city={Dublin},
                country={Ireland}}

\author[1]{Tahar Kechadi}[orcid=0000-0002-0176-6281]
\ead{tahar.kechadi@ucd.ie}
\ead{second.author@email.com}
             
\author[1]{Nhien-An Le-Khac}[orcid=0000-0003-4373-2212]
\cormark[1]
\ead{an.lekhac@ucd.ie}

\cortext[cor1]{Corresponding author}

\begin{abstract}
\doublespacing
\noindent 
In recent years, data  science has evolved significantly. Data analysis  and mining processes become
routines in  all sectors of the  economy where datasets  are available. Vast data  repositories have
been  collected,  curated,  stored,  and  used  for  extracting  knowledge.  And  this  is  becoming
commonplace. Subsequently, we extract a large amount  of knowledge, either directly from the data or
through experts in  the given domain. The challenge now  is how to exploit all this  large amount of
knowledge that is previously known for  efficient decision-making processes. Until recently, much of
the knowledge gained  through a number of years  of research is stored in static  knowledge bases or
ontologies,  while more  diverse and  dynamic knowledge  acquired from  data mining  studies is  not
centrally and consistently managed. In this research, we propose a novel model called ontology-based
knowledge map  to represent  and store the  results (knowledge)  of data mining  in crop  farming to
build, maintain, and enrich  the process of knowledge discovery. The proposed  model consists of six
main sets:  concepts, attributes, relations, transformations,  instances, and states. This  model is
dynamic and  facilitates the access, updates,  and exploitation of  the knowledge at any  time. This
paper also proposes an architecture for handling this knowledge-based model. The system architecture
includes knowledge modelling, extraction, assessment,  publishing, and exploitation. This system has
been implemented and used in agriculture for crop management and monitoring. It is proven to be very
effective and promising for its extension to other domains.
\end{abstract}



\maketitle

\section{Introduction}
\label{sec:intro}

In  this era  of  digital agriculture,  crop  farming can  take  advantage of  all  the advances  in
Information and Communication Technologies. The data collection becomes routine. This study collects
large amounts of data from different perspectives. Data science and machine learning are at the core
of  agricultural  data analyses  and  decision-making  processes.  In  recent years,  the  knowledge
discovered from the data  analysis processes is the most diverse and dynamic  in digital farming. In
other words, digital agriculture benefited significantly from data mining, data analytics or, in the
more  general term,  "data science".  Several  data-driven studies  have been  conducted on  several
agricultural elements,  including soil, weather,  crop yield,  disease, fertilisers, etc.,  with the
view to  derive models  that govern  the phenomenon behind  agricultural processes,  forecasting, or
optimising  the usage  of  resources.  Moreover, significant  farming  knowledge  also derived  from
farmers'  and agronomists'  experiences. These  can  be incorporated  into some  advanced models  to
increase the reliability and precision of digital agriculture. 

The application of data science to digital  agriculture enabled some studies which were not possible
in the past. These  include data construction, forecasting models and  validation of some hypotheses
about efficient farming  techniques. In this context, there are  several computational soil studies,
for  example,  building  datasets  of   soil  profiles  \cite{shangguan2013china},  monitoring  soil
characteristics under effects of other factors  and crop yield \cite{bishop2001comparison}, or using
soil characteristics to predict other soil characteristics \cite{wang2019comparison}. Another common
application of  knowledge mining in  agriculture is crop  yield prediction, for  example, predicting
yield  or  wheat   yield  based  on  soil  attributes,  weather   factors,  and  management  factors
\cite{aggelopoulou2011yield},   \cite{liu2001neural},   \cite{pantazi2016wheat}.  Finally,   another
important application  of knowledge mining is  disease prediction or protection  plan. For instance,
detecting nitrogen stresses  in the early crop growth stage  of corn fields \cite{maltas2013effect},
or detecting  and classifying sugar  beet diseases  \cite{karimi2006application} is crucial  for the
farmers.

Data mining or data analysis process has four typical tasks; clustering, classification, regression,
and associations.  Some of these are  heavily used. So  far, data classification and  regression are
widely   used  in   digital   agriculture.  Data   classification  is   used   to  detect   diseases
\cite{karimi2006application},     predict    crop     yield     (e.g.,     low,    medium,     high)
\cite{papageorgiou2013yield}. Regression techniques  are mainly used to find  correlations or models
between    crop    yields    and   other    different    attributes    \cite{aggelopoulou2011yield},
\cite{bishop2001comparison}. Although  the data  analysis process is  more or  less straightforward,
however, it faces  many challenges; the number of  input parameters is usually very  large, the data
quality and its variety, and the lack of  prior knowledge and tangible hypotheses. The latter can be
dealt with by incorporating some know-how in the data science process itself, however, this know-how
is not always available as an input. These known results or models are often published in scientific
articles  and  reports.  They  are  not  ready  to  be  incorporated  into  advanced  data  analysis
methodologies. They need to  be extracted from these scientific reports  and publications, they need
to be represented in a  form that they can be exploited, and they need to  be stored in a common and
unified format to facilitate their exploration and retrieval.

The main requirement of  such knowledge representation is that the mined  results are not consistent
among  different  sources  of  knowledge.   For  example,  one  can  have  two  different  knowledge
representations from two separate data mining studies used to predict farming conditions to maximise
the  crop  yield  of  winter  wheat.  The  concepts of  {\it  high  crop  yield}  in  two  knowledge
representations in  \cite{papageorgiou2013yield} and \cite{natarajan2016hybrid} are  different. They
depend on the way they were defined based on the datasets and their context.  Therefore, the results
may not  be consistent. This issue  can also occur with  input attributes. In addition,  the farming
knowledge represented as ontology is static, and it  is difficult to apply to various regions having
diverse farming  conditions. The  knowledge represented  as rules-based in  expert systems  does not
scale  well because  it  is  not easy  to  refine the  rules  and check  the  coherence  of all  the
rules. Finally,  the knowledge mined from  data mining processes  is dynamic and flexible,  so their
representations may  be different  among data  mining processes. For  instance, several  data mining
results are stored as rules while others as vectors or trained models.

This study extended the definition of the  Ontology-based Knowledge map (OAK) model described in the
\cite{ngo2020oak}  to include  more  new  entities, such  as  State,  Relation, Concept,  Attribute,
Transformation  (Data Transformation  and Computing  Algorithm), Instance,  and Lexicon  in the  OAK
model. The extended definition is much more robust and inclusive to model any types of knowledge. We
also provided a new definition of knowledge representation to include the results of any data mining
(DM) technique, such  as clustering, classification, and association rules  mining. These include DM
process, Fact Knowledge, and Analysis technique (Classification, Clustering, Regression, Association
Rule algorithms). This  makes it not only  easy to infer the  knowledge but also easy  to extract it
from external documents, such as journal papers.  The study extended the OAK architecture to include
six more modules; Knowledge Miner,  Knowledge Modelling, Knowledge Extraction, Knowledge Assessment,
Knowledge  Publishing, and  Knowledge Exploitation.  We provided  new evaluation,  verification, and
validation processes for the proposed model. We proposed a novel knowledge assessment within the OAK
model. This  study built a prototype  for a knowledge repository  that can hold up  to 500 knowledge
representations  extracted from  1,000 articles  and reports.  Finally, we  developed an  innovative
Knowledge Browser to identify knowledge by concepts and roles. This browser is also used to evaluate
the ability of the proposed model to handle knowledge. The contribution of this extended model is to
support  data scientists  and  agronomists in  discovering knowledge  and  representing their  mined
knowledge. 

The next section gives  an overview of knowledge concepts and how to  create knowledge in the domain
of  agriculture. Section  \ref{sec:OAK}  describes the  details  of the  OAK  model, which  includes
relevant  definitions,  architecture and  its  main  modules.  Section \ref{sec:Impl}  presents  the
implementation of  the main components  in the  OAK architecture, whereas  Section \ref{sec:ExpEval}
deals with experiments  and the evaluation of the  OAK model. Finally, the paper  gives a conclusion
and several future works in Section \ref{sec:Conclusion}.

\section{OAK - Ontology-based Knowledge Map Model}
\label{sec:OAK}

This study  proposes a  model for  representing knowledge  extracted from  data mining  and analysis
techniques. The  proposed model consists of  eight components: {\it knowledge  representation}, {\it
  ontology}, {\it  knowledge map model}, {\it  concept}, {\it transformation}, {\it  instance}, {\it
  state},  and {\it  relation}. Moreover,  it also  has two  more elements,  {\it lexicon}  and {\it
  hierarchy}, which  are structured factors of  the model. In  the following content, we  define the
model and its components. Let's define each of these components.

\setlength{\abovedisplayskip}{-12pt}\setlength{\belowdisplayskip}{0pt}

\begin{definition}[Domain Concept]
\label{def:CD}
  A Domain Concept set \(\mathbb{C_D}\) is a set of typical concepts used in domain \(\mathbb{D}\).

  \begin{equation}
    \label{eq:CD}
    \mathbb{C}_{D} = \{ c_D: \text{where, concept} c_D \text{ is used in domain } \mathbb{D} \} 
  \end{equation}
\end{definition}

For  example, in  a specific  domain,  typical concepts  in \(\mathbb{C_D}\)  can be  \textit{crop},
\textit{crop yield}, \textit{soil}, \textit{temperature}.  

\begin{definition}[Computing Concept]
\label{def:CDM}
  A Computing Concept set \(\mathbb{C_{DM}}\) is a  set of data mining concepts \(\mathbb{C_{DM}}\) to
  delegate to data mining tasks. 

  \begin{equation}
    \label{eq:CDM}
    \mathbb{C}_{DM} = \{ c_{DM}: \text{where, concept } c_{DM} \text{ is used in data mining} \} 
  \end{equation}
\end{definition}

\begin{definition}[Concept]
\label{def:C}
  A  Concept  set  \(\mathbb{C}\)  is  a   union  of  Domain  Concepts  \(\mathbb{C_D}\)  in  domain
  \(\mathbb{D}\) and Computing Concepts \(\mathbb{C_{DM}}\) to delegate to data mining tasks.  

  \begin{equation}
    \label{eq:C}
    \mathbb{C} = \mathbb{C_D} \cup \mathbb{C_{DM}}
  \end{equation}
\end{definition}

For example,   data  mining   concepts   in   \(\mathbb{C_{DM}}\)  can   be   \textit{clustering},
\textit{classification},  \textit{regression},  and \textit{association  rule}.  \(\mathbb{C_{DM}}\)
also has relevant  concepts, such as \textit{dataset}, \textit{evaluation}  and \textit{toolkit} to
represent related concepts of the mined knowledge items. 

\begin{definition}[Data Transformation]
\label{def:TD}
 A  Data Transformation  set \(\mathbb{T_D}\)  is a  set of  functions \(f(a)\),  which are  used to
 transform  the value  of concept  \(c\) in  \(\mathbb{C_D}\) from  range \(\mathbb{R}_x\)  to range
 \(\mathbb{R}_y\); where \(\mathbb{R}_x\) and \(\mathbb{R}_y\) are value ranges of concept \(c\). 

\begin{equation}
\label{eq:TD}
	\mathbb{T}_D = \{ f(c): \mathbb{R}_x \rightarrow \mathbb{R}_y\ ; \ c \in \mathbb{C_D} \} 
\end{equation}
\end{definition}

\begin{definition}[Computing Algorithm]
\label{def:TDM}
  A Computing Algorithm  set \(\mathbb{T_{DM}}\) is a set of  algorithms \(algorithm(c)\), which are
  used in the data mining tasks. 

\begin{equation}
\label{eq:TDM}
    \mathbb{T}_{DM}      =     \{      algorithm(c):     \mathbb{R}_{\{condition\}}      \rightarrow
    \mathbb{R}_{\{target\}}\ ; \ c \in \mathbb{C_{D}} \} 
\end{equation}
\end{definition}
where, \(\mathbb{R}_{\{condition\}}\) and \(\mathbb{R}_{\{target\}}\) are value ranges of concept \(c\).\\
For example, \(\mathbb{T_{DM}}\)  can be {\it Decision Tree}, {\it  Neural Networks} (algorithms for
{\it Classification}  or {\it Regression}), or  {\it Apriori}, {\it FP-growth}  (algorithms for {\it
  Association Rule}), etc. 

\begin{definition}[Transformation]
\label{def:T}
  A  Transformation set  \(\mathbb{T}\)  is a  union of  Data  Transformations \(\mathbb{T_D}\)  and
  Computing Algorithms \(\mathbb{T_{DM}}\). 

\begin{equation}
\label{eq:T}
    \mathbb{T} =\mathbb{T}_{D} \cup \mathbb{T}_{DM} 
\end{equation}
\end{definition}

For example,  the pH value  can be  from 0  to 14, therefore  the basis transformer  function is
itself (i.e. \(f(x) = x\), where x is  the pH attribute, and defined as {\it Transformation\_SoilPH}
in  the ontology  AgriComO, described  in  Section \ref{sec:AgriComO}).  Moreover, a  transformation
\(f(x)\) can convert a value to a new value in a different range, e.g. [0 .. 1] or to a new label in
the list with predefined Equation \ref{eq:TD} (defined as {\it Transformation\_SoilPH\_Tier5} in the
ontology, used in \cite{ngo2019predicting}): 

\begin{eqnarray}
\setlength{\jot}{-2pt}
    f(x) = \left\{
    \begin{array}{ll}
     \text{"Strongly acidic"}    &  pH  \leqslant 5 \\
     \text{"Acidic"}             &  5 < pH < 7 \\
     \text{"Neutral"}            & pH = 7 \ \ \ \ \\
     \text{"Alkaline"}           &  7 < pH \leqslant 10 \\
     \text{"Strongly alkaline"} &  10 < pH \\
    \end{array}\right.
\end{eqnarray}

This transformation is  one of many functions of the  concept of soil pH. They can  be fixed if they
are commonly used in knowledge items, or they can be defined when in use. In general, if one concept
\(c\) is used to analyse and create knowledge, this concept will have one or more transformations. 

\begin{equation}
	c \in \mathbb{C_D}, \exists t \in \mathbb{T_D}:\ c \overset{hasTransformation}{\rightarrow} t
\end{equation}

\begin{definition}[Domain Instance]
\label{def:ID}
A Domain Instance set  \(\mathbb{I_D}\) is a set of instances \(i\),  which represents an individual
of a domain concept \(c\) in a knowledge representation. 

\begin{equation}
\label{eq:ID}
    \mathbb{I_{D}}=\{ i: \exists c \in \mathbb{C_{D}};\ i \overset{isA}{\rightarrow} c \}
\end{equation}
\end{definition}

For example,  {\it SoilPH\_006}  and {\it Yield\_006}  are instances of  concept {\it  SoilPH}, {\it
  Yield} respectively. 

\begin{definition}[Computing Instance]
\label{def:IDM}
  A Computing  Instance set  \(\mathbb{I_{DM}}\) is  a set  of instances  \(i\), which  represents an
  individual of a computing concept \(c\) in a knowledge representation. 

\begin{equation}
\label{eq:IDM}
    \mathbb{I_{DM}}=\{ i: \exists c \in \mathbb{C_{DM}};\ i \overset{isA}{\rightarrow} c \}
\end{equation}
\end{definition}

For example, {\it Classifier\_006}, {\it Cluster\_007},  {\it Dataset\_001} are instances of concept
{\it Classifier}, {\it Cluster}, {\it Dataset} respectively. 

\begin{definition}[Instance]
\label{def:I}
  An  Instance set  \(\mathbb{I}\) is  a union  of domain  instances \(\mathbb{I_D}\)  and computing
  instances \(\mathbb{I_{DM}}\).  

\begin{equation}
\label{eq:I}
    \mathbb{I} =\mathbb{I}_{D} \cup \mathbb{I}_{DM} 
\end{equation}
\end{definition}

\begin{definition}[State]
\label{def:S}
A State set  \(\mathbb{S}\) is a set of states  \(s\), which are real values of  instance \(i\) when
applying transformation \(t\): 

\begin{equation}
\label{eq:S}
    \mathbb{S}=\{ s : \exists i \in \mathbb{I}: i \overset{hasState}{\rightarrow} s \} 
\end{equation}
\end{definition}

So, if one concept \(c\) is used to analyse as well as create knowledge and its value is used in the
knowledge representation,  this concept will  have one or more  transformations and this  value will
belong to some instance. 

\begin{equation}
\label{eq:IhasT}
\begin{aligned}
 \forall s \in\mathbb{S}, \exists i \in \mathbb{I}, \exists c \in\mathbb{C},\exists t \in\mathbb{T}: 
                    & \ i \overset{isA}{\rightarrow} c , \\
                    & \ c \overset{hasTransformation}{\rightarrow} t, \\
                & \ i \overset{hasState}{\rightarrow} s \\
\end{aligned}
\end{equation}

For example, {\it SoilPH\_006} can have state 4.5  or 7.5 if the transformation of {\it SoilPH\_006}
is {\it Transformation\_SoilPH}. Moreover, {\it SoilPH\_006} can have state {\it "Strongly acidic"},
{\it  "Acidic"},   {\it  "Neutral"},  {\it  "Alkaline"},   or  {\it  "Strongly  alkaline"}   if  the
transformation    of    {\it    SoilPH\_006}   is    {\it    Transformation\_SoilPH\_Tier5}.    {\it
  Transformation\_SoilPH} and {\it Transformation\_SoilPH\_Tier5}  are two different transformations
as explained above.

\begin{definition}[Relation]
\label{def:R}
 The Relation set \(\mathbb{R}\) consists of a set of relations \(r\) between two concepts
 (\(c_1, c_2\));  two instances (\(i_1,  i_2\)), between concept  \(c\) and instance  \(i\), between
 concept \(c\) and transformation \(t\), or between instance \(i\) and state \(s\): 

\begin{equation}
\label{eq:R}
    \mathbb{R} = \{(c_1, r, c_2)\} \cup \{(i_1, r, i_2)\} \cup \{(c, r, i)\} \cup \{(c, r, t)\} \cup
    \{(i, r, s)\} 
\end{equation}
\end{definition}

The knowledge relation  set \(\mathbb{R}\) contains at least four  relation types: {\it subClassOf},
{\it isA},  {\it hasTransformation}, and  {\it hasState}. Relation  {\it subClassOf} is  between two
concepts, (\(c_1\),  {\it subClassOf},  \(c_2\)). Relation {\it  isA} is between  an instance  and a
concept, (\(i\),  {\it isA},  \(c\)). Relation {\it  hasTransformation} is between  a concept  and a
transformation  (\(c\), {\it  hasTransformation},  \(t\)).  Relation {\it  hasState}  is between  an
instance and a state  (\(i\), {\it hasState}, \(s\)). Moreover, the  relation set \(\mathbb{R}\) can
have many  other relation types,  which are  used to present  the relationships between  concepts or
between instances. 

\begin{definition}[Ontology]
\label{def:O}
 Ontology \(\mathbb{O}\)  is a set of  three elements, including concepts  \(\mathbb{C}\), relations
 \(\mathbb{R}\), and transformations \(\mathbb{T}\): 

\begin{equation}
\label{eq:O}
    \mathbb{O} = (\mathbb{C}, \mathbb{R}, \mathbb{T})
\end{equation}
\end{definition}

\begin{definition}[Knowledge Representation]
\label{def:KR}
 A  Knowledge Representation  set \(\mathbb{KR}\)  is a  set of  four elements,  including instances
 \(\mathbb{I}\),    relations   \(\mathbb{R}\),    transformations   \(\mathbb{T}\),    and   states
 \(\mathbb{S}\): 

\begin{equation}
\label{eq:KR}
    \mathbb{KR} = (\mathbb{I}, \mathbb{R}, \mathbb{T}, \mathbb{S})
\end{equation}

And each representation \(kr\) is an individual of \(\mathbb{KR}\):

\begin{equation}
    kr = (\{i\}, \{r\}, \{t\}, \{s\})
\end{equation}
\end{definition}

\begin{definition}[Knowledge Map Model]
\label{def:KM}
 Knowledge Map Model \(\mathbb{KM}\)  is a  set of  five elements  (\(\mathbb{C}\), \(\mathbb{I}\),
 \(\mathbb{R}\), \(\mathbb{T}\), and \(\mathbb{S}\)), which are corresponding sets of concept \(c\),
 instance \(i\), relation \(r\), transformation \(t\), and state \(s\). 

\begin{equation}
\label{eq:KM}
	\mathbb{KM} = (\mathbb{C}, \mathbb{I}, \mathbb{R}, \mathbb{T}, \mathbb{S})
\end{equation}
\end{definition}

\begin{definition}[Lexicon]
\label{def:L}
 The  Lexicon \(\mathbb{L}\)  consists of  a set  of terms  (lexical entries)  for five  elements of
 \(\mathbb{KM}\),  containing concepts  (\(\mathbb{L}_c\)), instances  (\(\mathbb{L}_i\)), relations
 (\(\mathbb{L}_r\)), transformations (\(\mathbb{L}_t\)), and  states (\(\mathbb{L}_s\)). Their union
 is the lexicon: 

\begin{equation}
\label{eq:L}
    \mathbb{L} = \mathbb{L}_c \cup \mathbb{L}_i \cup \mathbb{L}_r \cup \mathbb{L}_t \cup \mathbb{L}_s
\end{equation}
\end{definition}

\begin{definition}[Hierarchy]
\label{def:H}
 The Hierarchy \(\mathbb{H}\) of the KMaps is a concept tree, which consists of a set of concepts in
 the ontology and {\it subClassOf} relations. 

\begin{equation}
    \mathbb{H} = (\mathbb{C}', \mathbb{R}')
\end{equation}
\end{definition}
where, 

\begin{equation*}
\begin{aligned}
    \mathbb{C}' &  \subseteq \mathbb{C}  \text{, and  } \mathbb{R}'  & =  \{ (c_x,  subClassOf, c_y)
    \},\text{ } c_x, c_y \in \mathbb{C}' 
\end{aligned}
\end{equation*}

Based on Equation \ref{eq:R} and Equation \ref{eq:O}, the hierarchy is a subset of the ontology; $\mathbb{H} \subseteq \mathbb{O}$

In general, the Lexicon \(\mathbb{L}\) provides the vocabulary of the knowledge map \(\mathbb{KM}\),
while the hierarchy \(\mathbb{H}\) provides its hierarchical structure.

\subsection{Definitions for Knowledge Representation}
\label{sec:def4KRep}

\begin{definition}[Knowledge]
\label{def:K}
  A Knowledge  set (mined knowledge)  \(\mathbb{K}\) is  a set of  processes, which use  data mining
  functions (algorithms) and input conditions to predict or describe one or more output targets.

\end{definition}

  For example, using  {\it Linear Regression} and  weather conditions to predict  diseases of crops.
  Or, knowledge  is using {\it K-Mean},  weather conditions as well  as the status of  crop diseases
  from a given dataset to predict clusters of diseases.

\begin{equation}
    \mathbb{K} = \{ ( \{algorithm\}, \{conditions\}, \{target\} ) \}
\end{equation}
  
Knowledge has two detail levels, Process Knowledge  and Fact Knowledge. Process Knowledge is used to
represent knowledge  for processes,  while Fact  Knowledge is  used to  detail these  processes with
specific knowledge and specific process cases. 

\begin{equation}
\label{eq:K}
    \mathbb{K} = \mathbb{K_P} \cup \mathbb{K_F}
\end{equation}

\begin{definition}[Process Knowledge]
\label{def:KP}
A  Process Knowledge  set  \(\mathbb{K_P}\) is  a  set of  mined processes,  which  use data  mining
functions and input instances as conditions to predict one or more output attributes as targets.  

\begin{equation}
\label{eq:KProc}
\begin{aligned}
	\mathbb{K_P} = & \{ ( \{algorithm\}, \{i_{condition}\}, \{i_{target}\} ) \} \\
          where,   &\ algorithm \in \mathbb{T_{DM}}, i_{condition} \in \mathbb{I_D}, i_{target} \in \mathbb{I_D}
\end{aligned}
\end{equation}
\end{definition}

It means, the knowledge uses a set of computing algorithms (\(\{algorithm\} \in \mathbb{T_{DM}}\)), and a set of instances (\(\{i_{condition}\} \in \mathbb{C_{D}}\)) in domain \(\mathbb{D}\) as conditions to predict output instances (\(\{i_{target}\} \in \mathbb{I_{D}}\)) in domain \(\mathbb{D}\).

And, each process knowledge representation \(k_p\) is an individual of \(\mathbb{K_P}\). Based on Definition \ref{def:KR} Knowledge Representation, process knowledge representation \(k_p\) is defined as follows:

\begin{equation}
\setlength{\jot}{-2pt}
\begin{aligned}
    k_p = ( & \{i_{process}\} \cup \{i_{condition}\} \cup \{i_{target}\}, \\
            & \{r\}, \\
            & \{t_{algorithm}\} \cup \{t_{condition}\} \cup \{t_{target}\}, \\
            & \{\} ) \\
\end{aligned}
\end{equation}
\begin{equation*}
\begin{aligned}
    \text{where, }  r = & \{ (i_{process}, hasAlgorithm, t_{algorithm})\}\; \cup \\
                        & \{ (i_{process}, hasCondition, i_{condition})\}\; \cup \\
                        & \{ (i_{condition}, hasTransformation, t_{condition})\}\; \cup \\
                        & \{ (i_{process}, predicts, i_{target})\}\;        \cup \\
                        & \{ (i_{target}, hasTransformation, t_{target})\}\; \\
\end{aligned}
\end{equation*}
\normalsize

For example, knowledge model \(k\) is used to predict {\it CropYield} based on conditions of {\it Temperature}, {\it Rainfall}, and {\it SeedRate} by using {\it Principal Component Analysis} (PCA) algorithm (for classification). It is presented as follows:

\begin{equation}
\begin{aligned}
	k   = ( & \{Classifier: Algorithm\_PCA\}, \\
	        & \{Temperature\_006, Rainfall\_006, SeedRate\_006\}, \\
	        & \{Yield\_006\})
\end{aligned}
\end{equation}
\normalsize

where, \(Temperature\_006\), \(Rainfall\_006\), and \(SeedRate\_006\) represent condition concepts {\it Temperature}, {\it Rainfall}, and {\it SeedRate} respectively, while \(Yield\_006\) represent the output concepts of the knowledge.

\begin{definition}[Fact Knowledge]
\label{def:KF}
 A Fact Knowledge set \(\mathbb{K_F}\) is a set  of mined processes, which use data mining functions
 and input states  of instances as conditions to  predict one or more output states  of instances as
 targets. 

\begin{equation}
\label{eq:KFact}
\begin{aligned}
	\mathbb{K_F} = & \{ ( \{algorithm\}, \{s_{condition}\}, \{s_{target}\} ) \} \\
    \text{where, } &\ algorithm \in \mathbb{T_{DM}}, s_{condition} \in \mathbb{S}, s_{target} \in \mathbb{S}, \\
                   &\ \exists i_{condition} \in \mathbb{I_D}, i_{condition} \overset{hasState}{\rightarrow} s_{condition}, \\
                   &\ \exists i_{target} \in \mathbb{I_D}, \; \; \; \; \;  i_{target} \overset{hasState}{\rightarrow} s_{target}
\end{aligned}
\end{equation}
\end{definition}

It means, the knowledge item uses the set of computing algorithms (\(\{algorithm\} \in \mathbb{T_{DM}}\)), and a set of values (\(\{s_{condition}\}\), where \(\{s_{condition}\}\) belongs to \( \{i_{condition}\}, \) \(\{i_{condition}\} \in \mathbb{I_{D}}\)) of instances in domain \(\mathbb{D}\) as conditions to predict output values (\(\{s_{target}\}\), where, \(\{s_{target}\}\) belongs to \(\{i_{target}\}, \{i_{target}\} \in \mathbb{I_{D}}\)) of instances in domain \(\mathbb{D}\).

Moreover, as mentioned in Equation \ref{eq:IhasT}, if one concept \(c\) with its state \(s\) is used the knowledge, this concept will have one or more transformations. Therefore, the extension of Equation \ref{eq:KFact} is:

\begin{equation}
\label{eq:IhasS}
\begin{aligned}
	\mathbb{K_F} = & \{ ( \{algorithm\}, \{s_{condition}\}, \{s_{target}\} ) \} \\
\end{aligned}
\end{equation}

\footnotesize{
\begin{equation*}
\setlength{\jot}{-2pt}
\begin{aligned}
    \text{where, } algorithm \in   & \mathbb{T_{DM}}, s_{condition} \in \mathbb{S}, s_{target} \in \mathbb{S},\\
        \exists c_{condition} \in   & \mathbb{C_D}, \exists i_{condition} \in \mathbb{I_D}, \exists t_{target} \in \mathbb{T_D}: \\  
                   & i_{condition} \overset{hasTransformation}{\rightarrow} t_{condition}, \\ 
                   & i_{condition} \overset{isA}{\rightarrow} c_{condition}, \\
                   & i_{condition} \overset{hasState}{\rightarrow} s_{condition},\\
        \exists c_{target} \in   &  \mathbb{C_D}, \exists i_{target} \in \mathbb{I_D}, \exists t_{target} \in \mathbb{T_D}:\\ 
                   &  i_{target} \overset{hasTransformation}{\rightarrow} t_{target},\\
                   &  i_{target} \overset{isA}{\rightarrow} c_{target}, \\
                   &  i_{target} \overset{hasState}{\rightarrow} s_{target} \\
\end{aligned}
\end{equation*}
}
\normalsize

And, each fact knowledge representation \(k_f\) is a map individual of \(\mathbb{K_F}\). Based on Definition \ref{def:KR} Knowledge Representation, fact knowledge representation \(k_f\) is defined as follows:

\begin{equation}
\setlength{\jot}{-2pt}
\begin{aligned}
    k_f =( & \{i_{fact}\} \cup \{i_{condition}\} \cup \{i_{target}\}, \\
           & \{r\} , \\
           & \{t_{algorithm}\} \cup \{t_{condition}\} \cup \{t_{target}\}, \\
           & \{s_{condition}\} \cup \{s_{target}\} ) \\
\end{aligned}
\end{equation}
\footnotesize{
\begin{equation*}
\begin{aligned}
    \text{where, } r = & \{ (i_{fact}, hasAlgorithm, t_{algorithm})\}\; \cup \\
                        & \{ (i_{fact}, hasCondition, i_{condition})\}\; \cup \\
                        & \{ (i_{condition}, hasTransformation, t_{condition})\}\; \cup  \\
                        & \{ (i_{condition}, hasState, s_{condition})\}\; \cup \\
                        & \{ (i_{fact}, predicts, i_{target})\}\; \cup  \\
                        & \{ (i_{target}, hasTransformation, t_{target})\}\; \cup \\
                        & \{ (i_{target}, hasState, s_{target})\}\; \\
\end{aligned}
\end{equation*}}
\normalsize

For example, the knowledge item \(k\) uses the \textit{Principal Component Analysis} (PCA) algorithm to predict \textit{HighYield} based on the values of \textit{Temperature}, \textit{Rainfall}, and \textit{SeedRate} of 20\textdegree C, 100mm and 200, respectively. In which, HighYield is based on transformation \textit{Transformation\_Yield\_Tier3} to transform yield values.

\footnotesize{
\begin{equation}
\begin{aligned}
	k   = ( & \{Classifier: Algorithm\_PCA\}, \\
	        & \{Temperature\_006: Transformation\_Temperature, 20oC; \\
	        & \ Rainfall\_006: Transformation\_Rainfall, 100mm, \\
	        & \ SeedRate\_006: Transformation\_SeedRate, 200\ \}, \\
	        & \{Yield\_006: Transformation\_Yield\_Tier3, HighYield\})
\end{aligned}
\end{equation}
}
\normalsize

In general, different forms of mined knowledge, which can be represented in this model are divided into 4 types: \textit{Classification}, \textit{Regression}, \textit{Clustering}, and \textit{Association} as in the following definitions.

\begin{definition}[Classification] 
\label{def:Classification}
    Classification is a data mining function that assigns items in a collection to predefined target categories \cite{tan2006introduction}.
\end{definition}

It means a classification model \(k_{Classification}\) uses data mining algorithms (\(t\), \(t \in \mathbb{T}_{DM}\), and \(t\) is used for classification tasks) to assign items in a collection to a target label (\textit{target} of the model) based on their concept (\textit{conditions} of the model). Each model contains three main elements, classifier, conditions, and targets as in the following definition.

\begin{equation}
\begin{aligned}
    k_{Classification} = (& \{classifier:algorithm,state\}, \\
                          & \{condition:transformation,state\}, \\
                          & \{target:transformation,state\} )
\end{aligned}
\end{equation}

This classification model \(k_{Classification}\) is represented in the OAK model (as defined in Definition \ref{def:KR} Knowledge Representation) as below:

\footnotesize{
\begin{equation}
\setlength{\jot}{-2pt}
\label{eq:Kclass}
\begin{aligned}
    k_{Classification} = & (\{i\}, \{r\}, \{t\}, \{s\}) \\
    \{i\} = & \{i_{classifier} \} \cup \{i_{condition}\} \cup \{i_{target}\} \\
    \{r\} = & \{ (i_{classifier}, hasAlgorithm, t_{DM})\}\; \cup \\
            & \{ (i_{classifier}, hasCondition, i_{condition})\}\; \cup \\
            & \{ (i_{condition}, hasTransformation, t_{D})\}\; \cup \\
            & \{ (i_{classifier}, predicts, i_{target})\}\; \cup \\
            &  \{ (i_{target}, hasTransformation, t_{D})\}\; \\
    \{t\} = & \{ t_{DM}=algorithm(c) \in \mathbb{T}_{DM}, c=Classification)  \}\; \cup \\
            & \{ t_D=f(i): \mathbb{R}_x \rightarrow \mathbb{R}_y\ ; i \in \{i_{condition}\} \} \; \cup \\
            & \{ t_D=f(i): \mathbb{R}_x \rightarrow \mathbb{R}_y\ ; i \in \{i_{target}\} \} \\
    \{s\} = & \{ s \in \mathbb{S}, \exists i \in \{i_{condition}\}: i 
	\overset{hasState}{\rightarrow} s \} \cup \\
            & \{ s \in \mathbb{S}, \exists i \in \{i_{target}\}: i 
	\overset{hasState}{\rightarrow} s \} \\
\end{aligned}
\end{equation}
}
\normalsize

\begin{figure}[htbp]
    \centering
    \includegraphics[width=8cm]{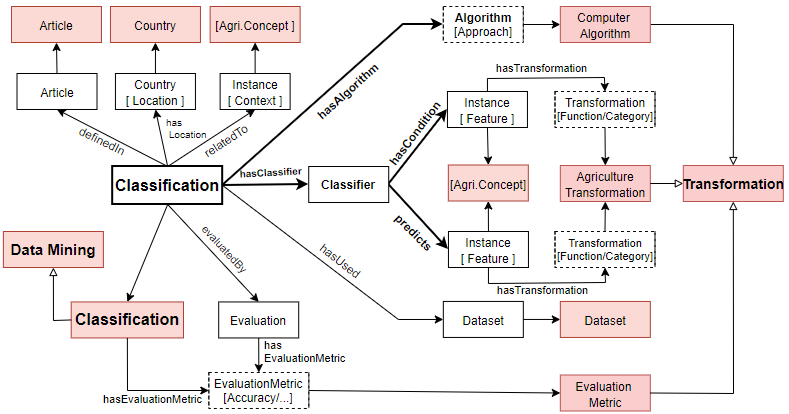}
    \caption{Visualisation of Knowledge Map Model for Classification}
    \label{fig:OKM:DMClassifyInOKM}
\end{figure}

This definition is a representation of classifications, which have the most important elements of knowledge, including algorithms, input conditions and output concepts. However, a classification model in data mining can have more detailed information, such as the context of experiments about location, the evaluation result, etc. For example, Figure \ref{fig:OKM:DMClassifyInOKM} shows that one classification can have more extra objects and several relationships linked to articles, countries, and other agriculture concepts (as environment).

\begin{definition}[Regression]
\label{def:Regression}
Regression is a data mining function that predicts a continuous outcome variable based on the value of condition variables.
\end{definition}

It means a regression model \(k_{Regression}\) uses data mining algorithms (\(t\), \(t \in \mathbb{T}_{DM}\), and \(t\) is used for regression tasks) to predict the value of a concept (\textit{target} of the model) based on their concepts (\textit{conditions} of the model).

\begin{equation}
\setlength{\jot}{-2pt}
\begin{aligned}
    k_{Regression} = (& \{regressor:algorithm,state\}, \\
                      & \{condition:transformation,state\}, \\
                      & \{target:transformation,state\} )
\end{aligned}
\end{equation}

Similarly, this regression model \(k_{Regression}\) is represented in the OAK model (as defined in Definition \ref{def:KR}) as follows:

\footnotesize{
\begin{equation}
\setlength{\jot}{-2pt}
\label{eq:Kregression}
\begin{aligned}
    k_{Regression} = & (\{i\}, \{r\}, \{t\}, \{s\}) \\
        \{i\} = & \{i_{regressor} \} \cup \{i_{condition}\} \cup \{i_{target}\} \\
        \{r\} = & \{ (i_{regressor}, hasAlgorithm, t_{DM})\}\; \cup \\
                & \{ (i_{regressor}, hasCondition, i_{condition})\}\; \cup \\
                & \{ (i_{condition}, hasTransformation, t_{D})\}\; \cup \\
                & \{ (i_{regressor}, predicts, i_{target})\}\; \cup \\
                &  \{ (i_{target}, hasTransformation, t_{D})\}\; \\
        \{t\} = & \{ t_{DM}=algorithm(c) \in \mathbb{T}_{DM}; c=Regression) \}\; \cup \\
                & \{ t_D=f(i): \mathbb{R}_x \rightarrow \mathbb{R}_y\ ; 
                  \ i \in \{i_{condition}\} \} \; \cup \\
                & \{ t_D=f(i): \mathbb{R}_x \rightarrow \mathbb{R}_y\ ; \ i \in \{i_{target}\} \} \\
        \{s\} = & \{ s \in \mathbb{S}, \exists i \in \{i_{condition}\}: i \overset{hasState}{\rightarrow} s \} \cup \\
                & \{ s \in \mathbb{S}, \exists i \in \{i_{target}\}:    i \overset{hasState}{\rightarrow} s \} \\
\end{aligned}
\end{equation}
}
\normalsize

\begin{definition}[Clustering]
\label{def:Clustering}
Clustering is a data mining function that finds clusters of data objects that are similar in some sense to one another \cite{tan2006introduction}. 
\end{definition} 

It means a clustering model \(k_{Clustering}\) uses data mining algorithms (\(t\), \(t \in \mathbb{T}_{DM}\), and \(t\) is used for Clustering tasks) to group similar items in a collection into clusters (\textit{cluster} of the model) based on their concepts (\textit{conditions} of the model).

\begin{equation}
\setlength{\jot}{-2pt}
\begin{aligned}
    k_{Clustering} = (& \{clustering:algorithm,state\}, \\
                      & \{condition:transformation,state\}, \\
                      & \{cluster:transformation,state\} )
\end{aligned}
\end{equation}
\normalsize

Similarly, this clustering model \(k_{Clustering}\) is represented in the OAK model (as defined in Definition \ref{def:KR}) as below:

\footnotesize{
\begin{equation}
\setlength{\jot}{-2pt}
\label{eq:Kcluster}
\begin{aligned}
    k_{Clustering} = & (\{i\}, \{r\}, \{t\}, \{s\}) \\
    \{i\} = & \{i_{clustering}\} \} \cup \{i_{condition}\} \cup \{i_{cluster}\} \\
    \{r\} = & \{ (i_{clustering}, hasAlgorithm, t_{DM})\}\; \cup \\
            & \{ (i_{clustering}, hasCondition, i_{condition})\}\; \cup \\
            & \{ (i_{clustering}, predicts, i_{cluster})\}\; \cup \\
            & \{ (i_{condition}, hasTransformation, t_{D})\}\;  \\
    \{t\} = & \{ t_{DM}=algorithm(c) \in \mathbb{T}_{DM}; c=Clustering)  \}\; \cup \\
            & \{ t_D=f(i): \mathbb{R}_x \rightarrow \mathbb{R}_y\ ; \ i \in \{i_{condition}\} \} \\
    \{s\} = & \{ s \in \mathbb{S}, \exists i \in \{i_{condition}\}: i 
	\overset{hasState}{\rightarrow} s \} \\
\end{aligned}
\end{equation}
}
\normalsize

\begin{definition}[Association Rule]
\label{def:Association}
Association Rule is a data mining function that discovers the probability of the co-occurrence of items in a collection. The relationships between co-occurring items are expressed as association rules \cite{tan2006introduction}.
\end{definition} 

It means an association model \(k_{Association}\) uses data mining algorithms (\(t\), \(t \in \mathbb{T}_{DM}\), and \(t\) is used for association tasks) to predict the co-occurrence of items (\textit{target} of the model) based on their concepts (\textit{conditions} of the model).

\begin{equation}
\setlength{\jot}{-2pt}
\begin{aligned}
    k_{Association} = (& \{association:algorithm,state\}, \\
                          & \{condition:transformation,state\}, \\
                          & \{rule:transformation,state\} )
\end{aligned}
\end{equation}

This association model \(k_{Association}\) is represented in the OAK model (as defined in Definition \ref{def:KR}) as below:

\footnotesize{
\begin{equation}
\setlength{\jot}{-2pt}
\label{eq:Kassociation}
\begin{aligned}
    k_{Association} = & (\{i\}, \{r\}, \{t\}, \{s\}) \\
    \{i\} = & \{i_{association}\} \cup \{i_{condition}\} \cup \{ i_{rule} \} \\
    \{r\} = & \{ (i_{association}, hasAlgorithm, t_{DM})\}\; \cup \\
            & \{ (i_{association}, hasCondition, i_{condition})\}\; \cup \\
            & \{ (i_{condition}, hasTransformation, t_{D})\}\;  \\
            & \{ (i_{association}, predicts, i_{rule})\}\; \cup \\
    \{t\} = & \{ t_{DM}=algorithm(c) \in \mathbb{T}_{DM}; c=Association)  \}\; \cup \\
            & \{ t_D=f(i): \mathbb{R}_x \rightarrow \mathbb{R}_y\ ; 
              \ i \in \{i_{condition}\} \} \; \cup \\
            & \{ t_D=f(i): \mathbb{R}_x \rightarrow \mathbb{R}_y\ ; \ i \in \{i_{target}\} \} \\
    \{s\} = & \{ s \in \mathbb{S}, \exists i \in \{i_{condition}\}: i 
	\overset{hasState}{\rightarrow} s \} \cup \\
            & \{ s \in \mathbb{S}, \exists i \in \{i_{target}\}: i 
	\overset{hasState}{\rightarrow} s \} \\
\end{aligned}
\end{equation}
}
\normalsize

\begin{figure}[ht]
	\centering
	\includegraphics[width=8cm]{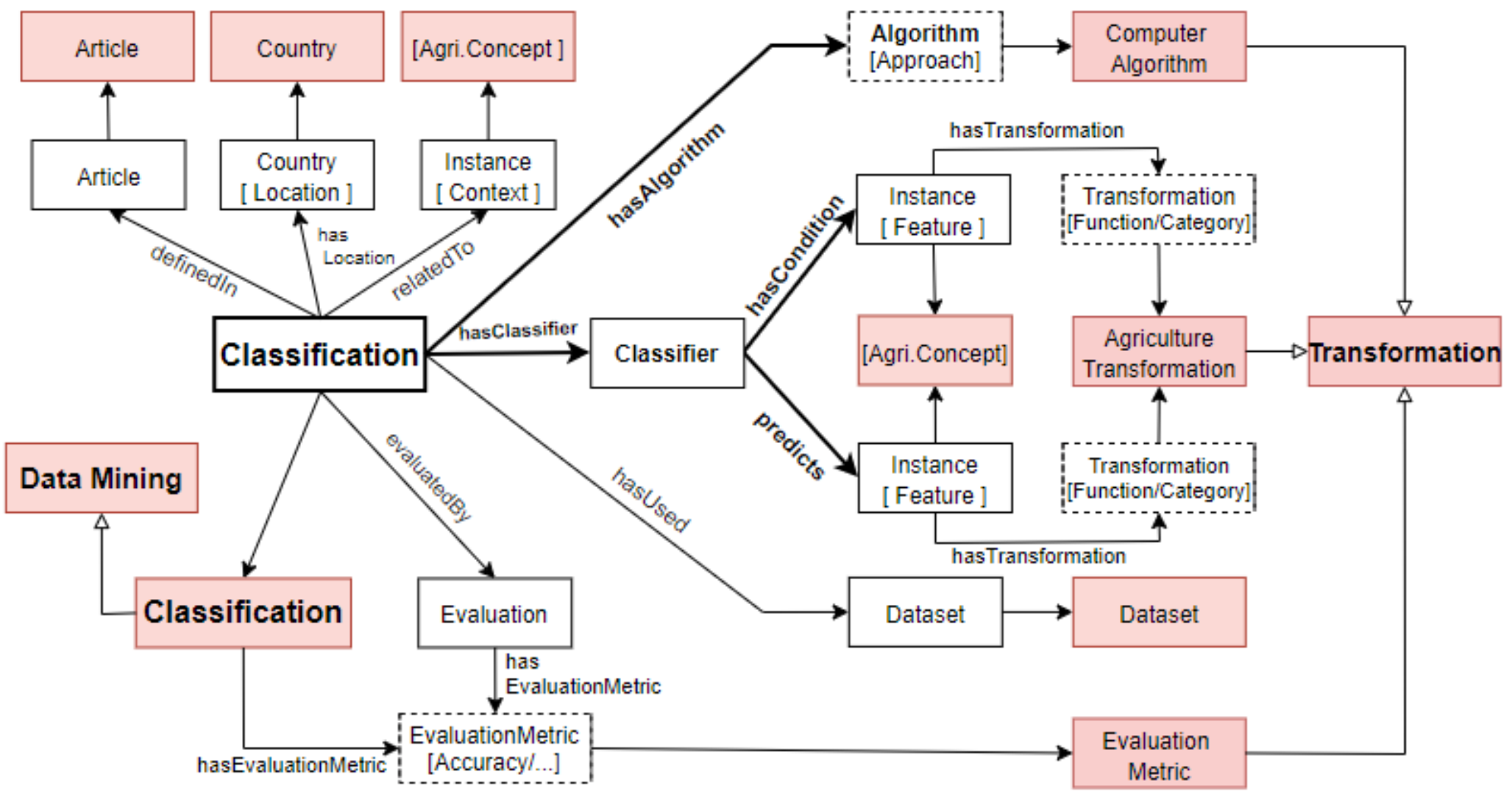}
	\caption{Main Data Mining Concepts in the Ontology.}
	\label{figDMconceptInOKM}
\end{figure}

\begin{definition}[Extend Knowledge]
\label{def:KE}
 Extend Knowledge is a mined process, which uses data mining functions to predict a target outcome based on condition states (as defined in Definition \ref{def:K}) and the context (containing context and locations of experiences), the dataset for training, and evaluation result. Its visualisation is shown in Figure \ref{figDMconceptInOKM}.
\end{definition}
\begin{equation}
\setlength{\jot}{-2pt}
\begin{aligned}
    k_{Extend} = ( & \{model:algorithm,state\}, \\
                   & \{condition:transformation,state\}, \\
                   & \{target:transformation,state\}, \\
                   & \{evaluation:evaluationmetric,state\},\\
                   & \{dataset\}, \{location\}, \{context\})
\end{aligned}
\end{equation}

This extend knowledge model \(k_{Extend}\) is represented in the OAK model (as defined in Definition \ref{def:KR}) as below:

\footnotesize{
\begin{equation}
\setlength{\jot}{-2pt}
\label{eq:KExtend}
\begin{aligned}
    k_{Extend} = ( & \{i_{extend}\} \cup \{i_{condition}\} \cup \{i_{target}\} \cup \\
                   & \{i_{dataset}\} \cup \{i_{location}\} \cup \{i_{context}\} \cup \{i_{evaluation}\}, \\
                   & \{r\} , \\
                   & \{t_{algorithm}\} \cup \{t_{condition}\} \cup \{t_{target}\} \cup \{t_{evaluationmetric}\}, \\
                   & \{s_{condition}\} \cup \{s_{target}\} \cup \{s_{evaluation}\} ) \\
\end{aligned}
\end{equation}
}
\footnotesize{
\begin{equation*}
\begin{aligned}
    \text{where,}\ r = & \{ (i_{extend}, hasAlgorithm, t_{algorithm})\}\; \cup \\
                        & \{ (i_{extend}, hasCondition, i_{condition})\}\; \cup \\
                        & \{ (i_{condition}, hasTransformation, t_{condition})\}\; \cup \\
                        & \{ (i_{condition}, hasState, s_{condition})\}\; \cup \\
                        & \{ (i_{extend}, predicts, i_{target})\}\; \cup \\
                        & \{ (i_{target}, hasTransformation, t_{target})\}\; \cup \\
                        & \{ (i_{target}, hasState, s_{target})\}\; \cup \\
                        & \{ (i_{extend}, hasDataset, i_{dataset})\}\; \cup \\
                        & \{ (i_{extend}, hasLocation, i_{location})\}\; \cup \\
                        & \{ (i_{extend}, hasContext, i_{context})\}\; \cup \\
                        & \{ (i_{extend}, hasEvaluation, i_{evaluation})\}\; \cup \\
                        & \{ (i_{evaluation}, hasEvaluationMetric, t_{evaluationmetric})\}\; \cup \\
                        & \{ (i_{evaluation}, hasState, s_{evaluation})\}\; \\
\end{aligned}
\end{equation*}}
\normalsize

\subsection{Architecture of Ontology-based Knowledge Map Model}
\label{sec:archiKMap}

\begin{figure*}[htbp]
	\centering
	\includegraphics[width=13cm]{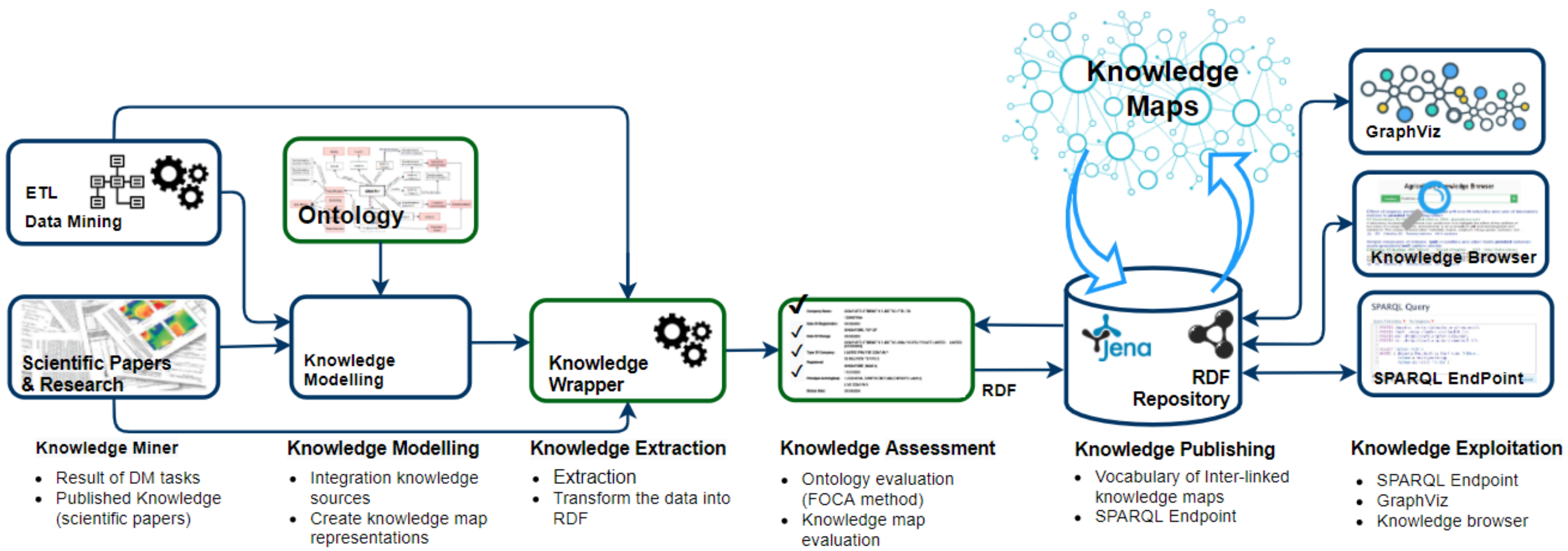}
	\caption{Architecture of Knowledge Map}
	\label{figKMArchitechture}
\end{figure*}

As illustrated in Figure \ref{figKMArchitechture}, the architecture of the OAK model consists of six components: (i) Knowledge Miner, (ii) Knowledge Modelling, (iii) Knowledge Extraction, (iv) Knowledge Assessment, (v) Knowledge Publishing and (vi) Knowledge Exploitation.

\begin{itemize} 
	\item \textbf{Knowledge Miner} is used to extract knowledge from data; this component can be a \textit{Data Mining} or publishing knowledge resource as \textit{Scientific Paper \& Research} module.
	\begin{itemize} 
		\item \textbf{Data Mining} refers to mining tools and techniques, which are used in analyzing datasets from various dimensions and perspectives, finding hidden knowledge and summarizing the identified relationships. These techniques are classification, clustering, regression, and association rules.
		\item \textbf{Scientific Papers} refers to publishing papers at scientific conferences or journals.
	\end{itemize}
	Depending on the source of the knowledge (from Data Mining tasks or from Scientific Papers) there are different ways to reproduce and represent that knowledge. If the knowledge is from the Data Mining module, they can be transformed using a rule-based module in the Knowledge Extraction module. Otherwise, if the knowledge is from scientific papers, the Knowledge extraction module uses natural language processing tools to extract them. 

	\item \textbf{Knowledge Modelling} is a select data mining pattern based on the data mining algorithm and generates the data mining instances (such as classification, clustering, regression, and association rule instances for the corresponding data mining tasks) and links to data mining algorithms as the transformation objects of the data mining instances. 

	\item \textbf{Knowledge Extraction} is the main module to transform the knowledge from the output of the knowledge miner module to the Knowledge Publishing module. This step is to identify concepts in mining results and locate them in the ontology. Basically, these concepts occur in the mined results as input features and predicting features, for example, \textit{SoilPH}, \textit{SeedRate}, \textit{Nitrogen}, \textit{Wheat}, and \textit{MeanYield}. If the knowledge source is from scientific papers, this module is based on natural language processing tools to extract published knowledge from the articles, for example, entity extraction \cite{ngo2021domain} or relation extraction \cite{luan2018multi}.

	\item \textbf{Knowledge Assessment} is the module to verify and rate the mined knowledge from the output of the knowledge miner module before importing them into the \textit{Knowledge Publishing} module to store them. This module evaluates knowledge representations based on their contributed parts and grades them on a scale of 100, in which any knowledge representations with less than 50 should not be imported to the system.

	\item \textbf{Knowledge Publishing} is a graph database server, which supports RDF triple storage and SPARQL protocol for retrieval. This module receives the domain knowledge from the pre-defined ontology, the mined knowledge representations from the \textit{Knowledge Wrapper} module, and then stores it in the RDF Triple Storage as a set of RDF turtles. 

	\item \textbf{Knowledge Exploitation} are application components, which are used to search and represent knowledge as required by users. 
\end{itemize} 

The  ontology-based knowledge  map model  includes  two knowledge  layers.  The first  layer is  the
background  knowledge  about agriculture,  defined  as a core  KMap  and  built from  a  pre-defined
agricultural ontology (mainly  cropping knowledge in this  project). This layer defines  most of the
concepts  (agricultural  entities related  to  crops)  in the  KMaps  and  common relations  between
them. The  second layer includes  knowledge representations of  data mining results.  This knowledge
layer is extracted from the data mining process and imported by the Knowledge Wrapper module.

\section{Implementation}
\label{sec:Impl}

To implement this empirical study and to realize the proposed model, this project is broken into four major phases including building an agriculture ontology, knowledge wrapper representing knowledge models from data mining results, and knowledge browser.

\subsection{Agriculture Computing Ontology}
\label{sec:AgriComO}

There are several ontologies for agriculture. Each ontology has a specific purpose in agriculture studies rather than using it to handle knowledge in digital agriculture. For example, AGROVOC\footnote{http://aims.fao.org/vest-registry/vocabularies/agrovoc} provides vocabularies in agriculture \cite{caracciolo2013agrovoc}, AgOnt aims to agriculture IoT \cite{hu2010agont}, Citrus Ontology focuses on citrus fruits \cite{wang2018citrus}, or Plant Ontology (PO)  describes plant anatomy, morphology and growth and development \cite{cooper2012plant}. Therefore, it is necessary to create a new ontology, which includes common agriculture concepts and their relationships with each other.

Basically, most ontologies describe classes (concepts), instances, attributes, and relations. Moreover, some ontologies also include restrictions, rules, axioms, and function terms. However, as a formal presentation of KMaps, we propose an ontology with the following components:

\begin{figure}[ht]
	\centering
	\includegraphics[width=8.2cm]{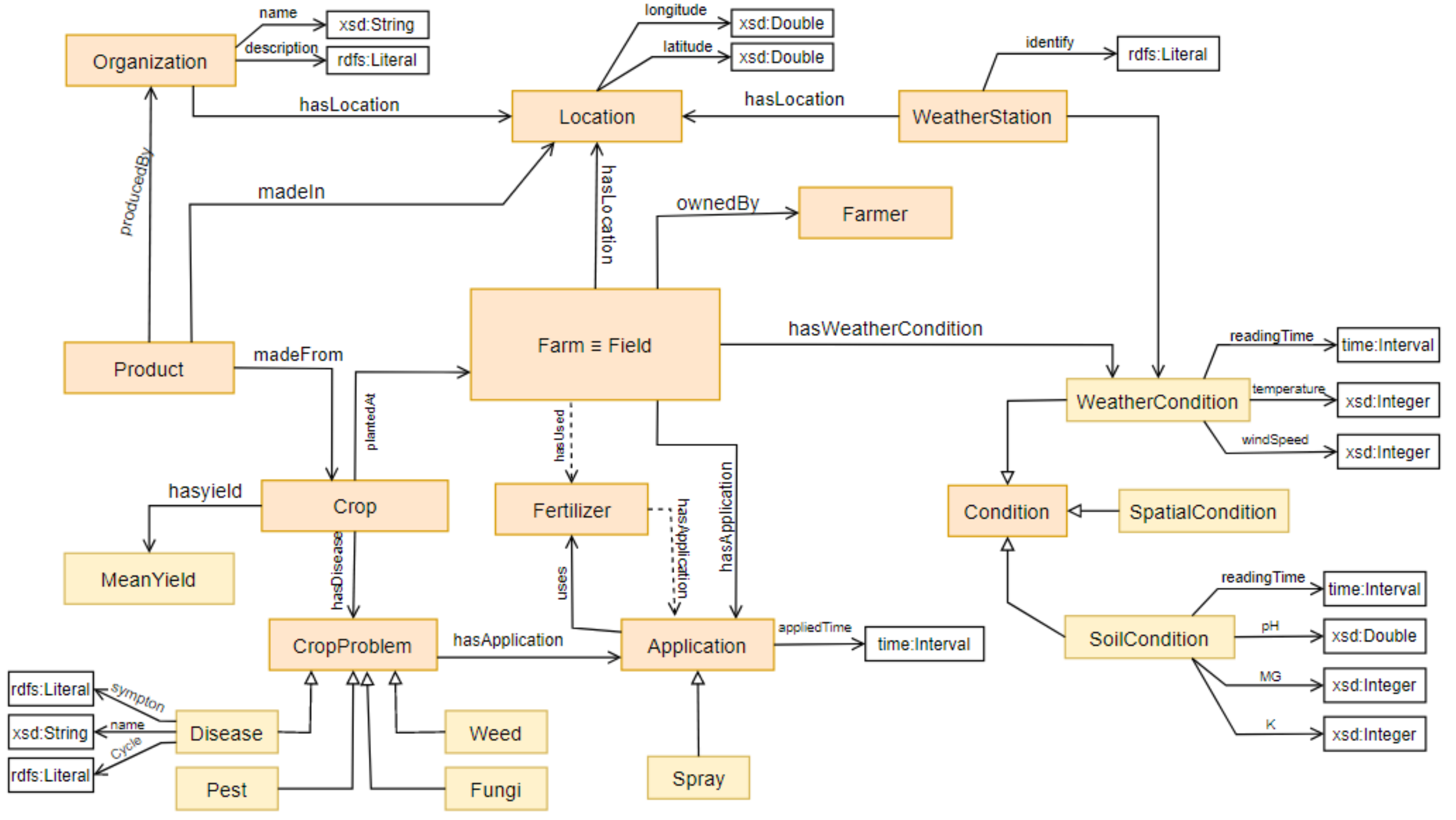}
	\caption{An overview of Agricultural Ontology Architecture}
	\label{figAgriOntology}
\end{figure}

\begin{itemize} 
	\item \textbf{Concepts}: Concepts in the ontology include concepts in agriculture and concepts for representing four main tasks of data mining. For example, agriculture concepts have \textit{field}, \textit{farmer}, \textit{crop}, \textit{organization}, \textit{location}, and \textit{product}, while data mining concepts have \textit{clustering}, \textit{classification}, \textit{regression}, and \textit{association rule}.
	\item \textbf{Transformations}: They are pre-defined transformation functions of agriculture concepts and existing data mining techniques for four main tasks of data mining.
		\begin{itemize} 
			\item Agriculture Transformation
			\item Agroclimatic Indices
			\item Spectral Vegetation Indices
			\item Computer Algorithms
			\item Evaluation Metrics
		\end{itemize} 
    \item \textbf{Relations}: ways in which concepts (and then instances) can be related to others. They are defined as the \(\mathbb{R}\) set in 		definitions. 
\end{itemize} 

In this research phase, we propose an agricultural ontology for the purpose of using it in the knowledge map model. The agricultural ontology contains 4 sub-domains: agriculture, IoT, geography, and the business sub-domain (Figure \ref{figAgriOntology}). In addition, the ontology is also added concepts in the data mining domain as shown in Figure \ref{figDMconceptInOKM}. These concepts and relations will be knowledge frameworks to transform mined knowledge from data mining tasks to the knowledge representations and import them into the knowledge maps.

\begin{table}[ht]
\selectfont\scriptsize\centering
    \renewcommand{\arraystretch}{1.1}
	\caption{AgriComO's Ontology Metrics}
	\label{tableOntologyMetrics}%
	\setlength{\tabcolsep}{2pt}
    \begin{tabular} {lrrrr}
    \hline\noalign{\smallskip}
        & & \textbf{with} & \textbf{with} & \textbf{with} \\
            \textbf{Figure} & \hspace{0.5cm}\textbf{Core} & 
            \hspace{0.2cm}\textbf{Transformation} & 
            \hspace{0.2cm}\textbf{Geo-data}   & 
            \hspace{0.2cm}\textbf{Disease}   \\
    \hline\noalign{\smallskip}
        \textbf{Axiom}  	        & 8,319 & 17,387 & 92,046 & 145,732 \\
        \textbf{Logical axiom} 	    & 3,794 &  7,159 & 55,503 &  86,355 \\
        \textbf{Declaration axioms} & 1,780 &  3,091 &  8,294 &  15,409 \\
        \textbf{Class} 	            &   606 &    606 &    606 &     606 \\
        \textbf{Object property}    &   126 &    126 &    126 &     126 \\
        \textbf{Data property}      &   164 &    164 &    164 &     164 \\
        \textbf{Individual}         &   864 &  2,175 &  7,378 &  14,495 \\
    \hline\noalign{\smallskip}
        \textbf{Triple count}       & 8,369 & 17,437 & 92,096 & 145,786 \\
    \hline\noalign{\smallskip}
    \end{tabular}%
\end{table}%

After building a  knowledge hierarchy, the ontology  provides an overview of  the agriculture domain
and describes agricultural concepts, life  cycles between seeds, plants, harvesting, transportation,
and consumption.  It also gives  relationships between  agricultural concepts and  related concepts,
such as  weather, soil conditions,  fertilizers, and farm  descriptions. In addition,  this ontology
also includes data mining concepts, such  as classification, clustering, regression, and association
rule. Combined with  agricultural concepts, these are used to  represent mined knowledge. Currently,
the ontology  has 598  classes and  over 18,176  axioms related  to agriculture  (as shown  in Table
\ref{tableOntologyMetrics}, and partly presented in \cite{ngo2018ontology}). It provides an overview
of the  agriculture domain with the  most general agricultural  concepts. As a result,  the AgriComO
ontology can  be used as the  core ontology to build  the knowledge maps for  agriculture. Moreover,
this ontology with agricultural hierarchy can help  to integrate available resources to build larger
and more precise knowledge maps in the agriculture domain.

\subsection{Knowledge Wrapper}
\label{sec:buildPhase2}

The procedure for mapping mined knowledge into a knowledge representation in the ontology-based knowledge map is defined in the \textit{Knowledge Wrapper} module. The knowledge wrapper module is the main module to transform the mined knowledge into a knowledge representation \(k\) (as defined by \(k = (\{i\}, \{t\}, \{s\}, \{r\})\) in \textit{Definition \ref{def:KR}}, Section \ref{sec:OAK}). Then, it is converted into RDF turtles and imported into the RDF Triple storage. 

This module has six steps:

\begin{itemize}
	\item \textbf{Step 1, Identify model}: Select data mining pattern based on the data mining algorithm and generate the data mining instances (such as classification, clustering, clustering, and association rule instances for the corresponding data mining tasks, as defined in Definition \ref{def:Classification}-\ref{def:Association}) and link to data mining algorithm as the transformation objects of the data mining instances (as defined in Definition \ref{def:TDM}).
	\item \textbf{Step 2, Identify concepts}: Identify agricultural concepts in mining results and locate them in the ontology. Basically, these concepts occur in the mined results as input features and predicting features, for example, \textit{SoilPH}, \textit{SeedRate}, \textit{Nitrogen}, \textit{Wheat}, \textit{MeanYield}.
	\item \textbf{Step 3, Generate instances}: Generate agricultural instances of each located concept and link them to data mining instances (in \textbf{Step 1}) based on the framework of data mining tasks.
	\item \textbf{Step 4, Identify transformations}: Identify transformations of each concept in the mining results (as defined in Definition \ref{def:TD}), locate them in the ontology part of the KMap, then link them to agricultural instances (in \textbf{Step 3}). 
	\item \textbf{Step 5, Generate states}: Identify states of each concept in the mining results, generate states and link to instances (in \textbf{Step 3}) in the knowledge representation.
	\item \textbf{Step 6, Generate turtles}: Transform the knowledge representation into RDF turtles and import them into the RDF triple storage.
\end{itemize}

The set of instances \(\{i\}\) is created in Step 1 and Step 3, while the set of transformations \(\{i\}\) is created in Step 1 and Step 4. The set of states \(\{s\}\) is generated in Step 5, however, not all knowledge representations have sets of states. For example, in the model to predict crop yield, the input values only occur when the model is executed. Therefore, the set of states for this knowledge representation is nearly none. Finally, set of relations \(\{i\}\) is based on relation \textit{isA}, \textit{hasTransformation}, \textit{hasState}, \textit{hasCondition}, and \textit{predicts}.

As a result, an example of \textit{Classifier\_010} has shown that this knowledge item is a classifier and it is published in the article \textit{Article\_010} \cite{pantazi2016wheat}. This classifier uses cation exchange capacity, organic carbon, soil attributes (calcium, magnesium, nitrogen, soil moisture, and soil pH) to predict crop yield for wheat based on Counter-propagation Artificial Neural Network (CPANN), Supervised Kohonen Network (SKN), and XY-fusion network (XYF) algorithms. This knowledge has been conducted in the United Kingdom (Listing \ref{lstClassifer010}).

\begin{lstlisting}[label=lstClassifer010,caption=Triples of Classifier\_010]
AgriKMaps:Classifier_010 
    rdf:type owl:NamedIndividual ,
             AgriComO:Classifier ,
             AgriComO:KnowledgeModel ;
    rdfs:label "Classifier 010" .
    AgriComO:definedIn 
            AgriKMaps:Article_010 ;
    AgriComO:hasAlgorithm 
            AgriComO:Algorithm_CPANN ,
            AgriComO:Algorithm_SKN ,
            AgriComO:Algorithm_XYF ;
    AgriComO:hasCondition 
            AgriKMaps:CEC_010 ,
            AgriKMaps:OrganicCarbon_010 ,
            AgriKMaps:SoilCa_010 ,
            AgriKMaps:SoilMG_010 ,
            AgriKMaps:SoilMoisture_010 ,
            AgriKMaps:SoilN_010 ,
            AgriKMaps:SoilPH_010 ;
    AgriComO:predicts 
            AgriKMaps:Yield_010 ;
    AgriComO:evaluatedBy 
            AgriKMaps:Evaluation_010 ;
    AgriComO:hasLocation 
            AgriComO:United_Kingdom ;
    AgriComO:relatedTo AgriComO:Wheat ;
    AgriComO:grade 60 ;
\end{lstlisting}

\subsection{Knowledge Map Assessment}

This section describes the knowledge assessment progress and result of the mined knowledge of agriculture in Section \ref{sec:KnowMaterials}. We divide information in each knowledge map representation into three groups: Basic information, principal information and subordinal information. The rates of the three groups are 20\%, 40\% and 40\% for basic, principal, and subordinal information, relatively, as below.

\begin{itemize}
	\item \textbf{Basic information} (20\%): general information of the knowledge, such as authors, title or time of research, resource of knowledge.
	\item \textbf{Principal information} (40\%): major information of knowledge, such as algorithms, conditions, target, transformations for conditions and target.
	\item \textbf{Subordinal information} (40\%): dataset, evaluation (metrics and values), locations and context of research.
\end{itemize}

In general, each knowledge representation will be rated around 50-60\% if it has basic information for basic knowledge as in \textit{Definition \ref{def:K}}. However, its grade can reach 100\% if it has full information for extending knowledge as in \textit{Definition \ref{def:KE}}.

\begin{table}[ht]
\centering
	\caption{Knowledge Map Assessment}
	\label{tableKnowledgeAssessment}%
    \begin{tabular} {lr}
    \hline\noalign{\smallskip}
        \textbf{Figure} & \textbf{Count/Rate} \\
    \hline\noalign{\smallskip}
        Scientific article       & 500  \\
        Knowledge representation & 500  \\
    \hline\noalign{\smallskip}
        Basic information 	     & 100\%  \\
        Principal information 	 & 100\%  \\
        Subordinal information 	 & 65.2\% \\
    \hline\noalign{\smallskip}
        Knowledge map rate 	     & 76.5\%\\
    \hline\noalign{\smallskip}
    \end{tabular}%
\end{table}%

Table \ref{tableKnowledgeAssessment} shows the result of knowledge assessment based on 500 mined knowledge items, which are extracted from the knowledge resources as shown in Section \ref{sec:KnowMaterials}.

\subsection{Knowledge Publishing}

We propose to use the native knowledge storage with the Triple store technology. Literally, \textbf{Apache Jena}\footnote{https://jena.apache.org/index.html} is a native knowledge graph storage technology. It is used for SPARQL Engine with \textbf{Fuseki}\footnote{https://jena.apache.org/documentation/fuseki2/} for SPARQL Endpoint.

In this module, both ontology and knowledge maps are transformed into RDF format. Concepts in AgriComO\footnote{Prefix of AgriComO: \textit{http://www.ucd.ie/consus/AgriComO\#}} and instances in AgriKMaps\footnote{Prefix of AgriKMaps: \textit{http://www.ucd.ie/consus/AgriKMaps\#}} have IRIs to identify. In addition, common concepts and relationships in AgriComO are considered to refer to elements of the DCMI\footnote{https://www.dublincore.org/specifications/dublin-core/dcmi-terms/} (Dublin Core Metadata Initiative) for further sharing.

In the context of a management system, the query language for knowledge retrieval is SPARQL 1.1 Query Language\footnote{https://www.w3.org/TR/sparql11-query/}. Apache Jena also provides a REST API for the Knowledge Wrapper module, which can access to query and import knowledge representations. Moreover, Apache Jena also provides HTTP and the SPARQL protocols for directly queries on web pages. The SPARQL Endpoint is installed locally and can be accessed for web-based queries. 
 
\begin{figure}[ht]
    \centering
    \includegraphics[width=8.5cm]{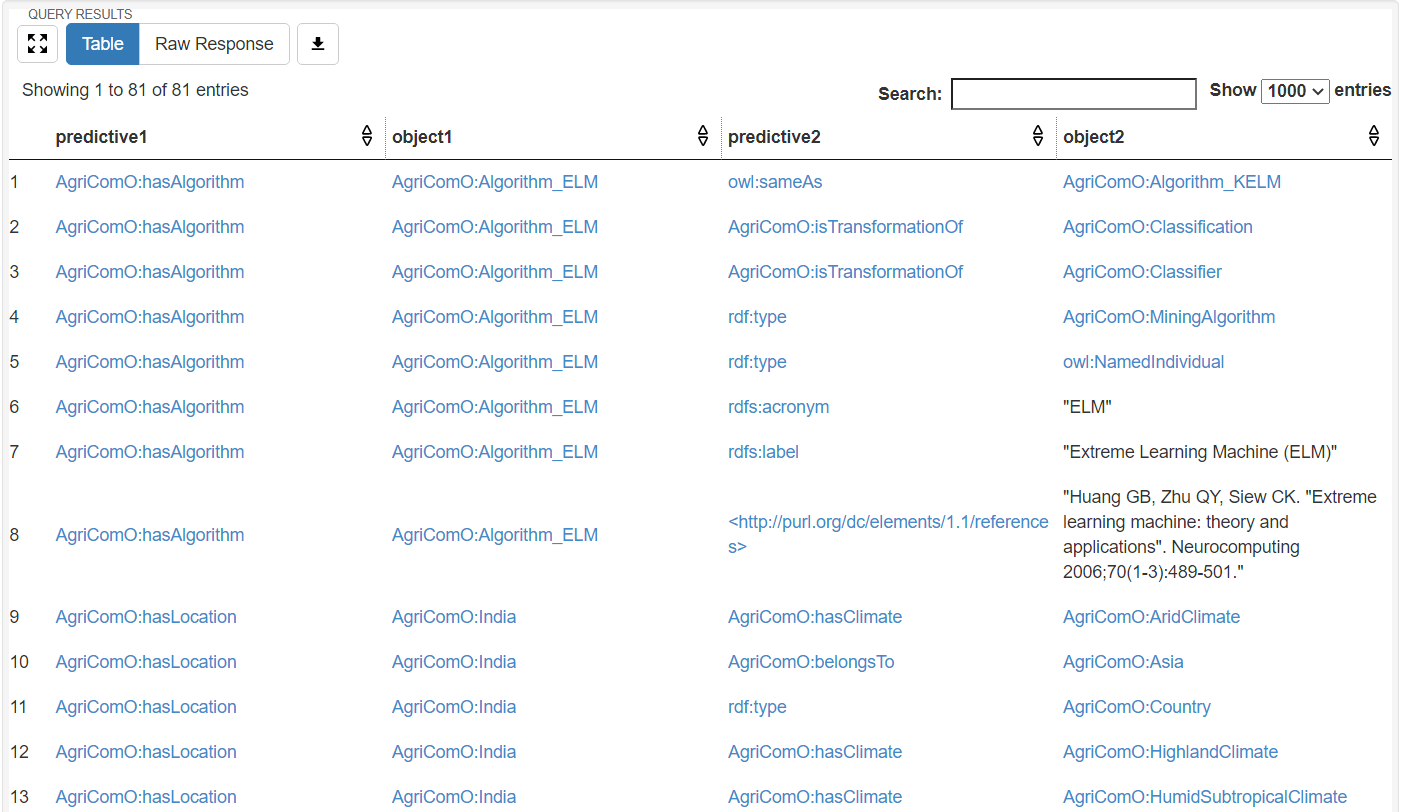}
    \caption{Example of SPARQL Query on AgriKMaps.}
    \label{figSPARQLsample}
\end{figure}

Specifically, SPARQL queries can be run on the SPARQL Endpoint with a web interface. For example, Figure \ref{figSPARQLsample} shows the results for the query "What are conditions and the target attribute of the knowledge model \(Regressor\_004\)?". The SPARQL query for this question is shown below:

\begin{lstlisting}[label=lstRegressor0001, caption=SPARQL Query details for Regressor\_0001]
PREFIX rdf: <http://www.w3.org/1999/02/22-rdf-syntax-ns#> 
PREFIX rdfs: <http://www.w3.org/2000/01/rdf-schema#> 
PREFIX owl: <http://www.w3.org/2002/07/owl#> 
PREFIX AgriComO: <http://www.ucd.ie/consus/AgriComO#> 
PREFIX AgriKMaps:<http://www.ucd.ie/consus/AgriKMaps#> 
SELECT *
WHERE {
     AgriKMaps:Regressor_0001 ?predictive1 ?object1 .
     ?object1               ?predictive2 ?object2
}
\end{lstlisting}

\subsection{Knowledge Browser}

\begin{figure}[ht]
    \centering
    \includegraphics[width=8.5cm]{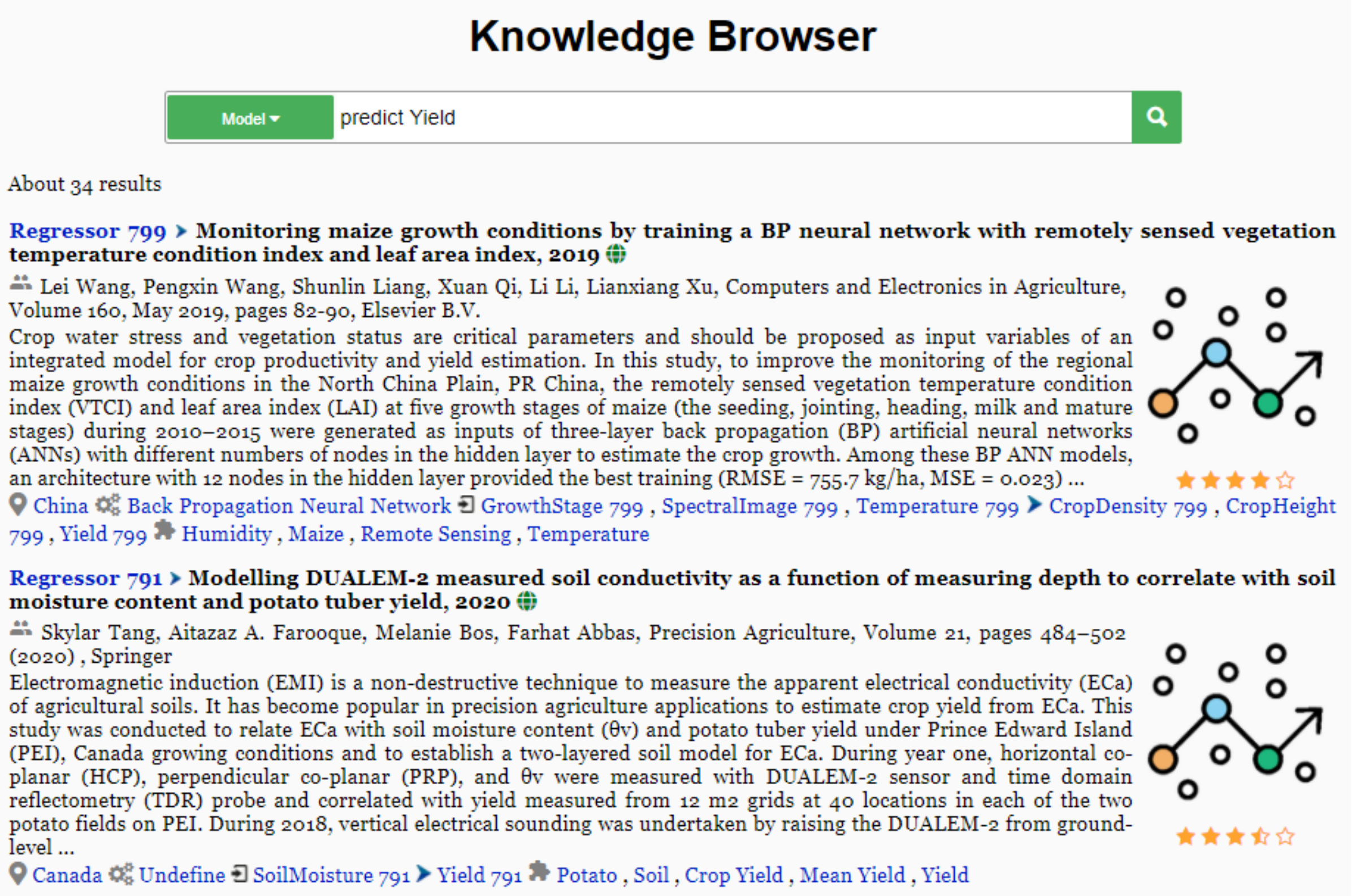}
    \caption{Screenshot of Knowledge Browser}
    \label{figAgriKnowledgeBrowser}
\end{figure}

\textbf{Knowledge Browser} is one of the knowledge exploitation applications. Knowledge Browser helps users to find mined knowledge items based on their input queries or keywords. Basically, the input queries can be a simple sentence, such as "predict crop yield in the United Kingdom" (with the result as shown in Figure \ref{figAgriKnowledgeBrowser}). The process includes:

\begin{itemize}
    \item Finding concepts from search queries, 
    \item Segmenting concepts into parts knowledge models
    \item Generating SPARQL queries
    \item Regenerating result from return triples.
\end{itemize}

\section{Experimental Results}
\label{sec:ExpEval}

Overall, there is no approach to evaluate the whole model in knowledge management. In this study, each individual component of the proposed OAK model has been evaluated. Firstly, this project applied several approaches to validate and verify the proposed AgriComO ontology \cite{bandeira2016foca}. Next, the project also completed several experiments on the knowledge management system to demonstrate the ability of the OAK model in knowledge management and handling mined knowledge in digital agriculture.

\subsection{Knowledge Materials}
\label{sec:KnowMaterials}

Materials for mined knowledge representations are extracted from scientific papers published in two journals in the digital agriculture domain. The total number of the articles is 3,381 articles and it is filtered by a keyword list to select about 1,000 papers, which present data mining results (as shown in Table \ref{tableMaterials}). The keyword list for filtering includes 10 keywords: agriculture, crop, wheat, oats, rice, fertiliser, soil, weather, nutrient, disease, data mining, and machine learning.
\begin{table}[ht]
\centering
  \caption{Materials for Agriculture Knowledge Maps}
  \label{tableMaterials}%
    \begin{tabular} {lr}
    \hline\noalign{\smallskip}
        \textbf{Figure} & \textbf{Count}     \\
    \hline\noalign{\smallskip}
        \textbf{Scientific Article} & 3,381  \\
        \textbf{Crop Articles} 	    & 1,972  \\
        \textbf{Data mining}        & 1,007  \\
        \textbf{Classification}     &   467	 \\
        \textbf{Clustering} 	    &   249	 \\
        \textbf{Regression}         &   189  \\
        \textbf{Association Rule}   &    95	 \\
    \hline\noalign{\smallskip}
    \end{tabular}%
\end{table}%

Each article is considered to extract 9 elements of knowledge representation (as listed in Table \ref{tableKnowledgeFeatures}). Basically, there are several parts that are extracted semi-automatically while other parts are manually extracted from the content of the article by using the \textit{Knowledge Wrapper} module. As result, there are 500 knowledge representations extracted from the dataset and they are imported into the \textit{Knowledge Publishing} module as knowledge maps of agriculture.

\begin{table}[ht]
\selectfont\centering
  \caption{Knowledge Features of each Knowledge Representation}
  \label{tableKnowledgeFeatures}%
    \begin{tabular}{|m{2.3cm}|m{5.5cm}|}
    \hline
        \textbf{Figure} & \textbf{Description} \\
    \hline
    	\textbf{Knowledge task} & Extract the clustering, classification, regression or association type from the title. \\
    	\textbf{Approaches} 	& Extract mining algorithms from abstract. \\
    	\textbf{Conditions} 	& Manually extract input attributes from the main section of the model.\\
    	\textbf{Target} 		& Extract output attributes, predict label, or rule sets from the title. \\
    	\textbf{Transformation} & Manually extract Transformations formulas of attributes.\\
    	\textbf{Evaluation} 	& Extract from abstract or section Experience, refer to \textit{Evaluation Metrics} in \textit{Transformations}.\\
    	\textbf{Dataset} 		& Extract information of dataset, the data size of experiences	from section Material or Experience. \\
    	\textbf{Location} 		& Extract locations of experiences from section Material.\\
    	\textbf{Context} 		& Context of study and mined knowledge Manually extract from section Material. \\
    \hline
    \end{tabular}%
\end{table}%

\subsection{Agriculture Ontology Evaluation}
\label{sec:OntoEval}

Although ontologies are formal representations of knowledge and are widely used in computer science,
there is no standard unity of benchmarks or metrics to evaluate them. However, there is a remarkable
number of  studies focused on  evaluating them. Brank  et al. \cite{brank2005survey}  summarised six
levels for  ontology evaluation. They are  (i) Lexical, vocabulary; (ii)  Hierarchy, taxonomy; (iii)
Semantic relation; (iv) Concept, application; (v) Syntactic level; and (vi) Structure, design. There
are several  approaches for evaluating ontologies  on different levels, such  as lexical, vocabulary
\cite{maedche2002measuring},  hierarchy,  taxonomy \cite{dellschaft2008strategies,  porzel2004task},
and semantic relation \cite{porzel2004task}. These evaluation approaches are based on application or
humans, while the study of Staab et  al. \cite{maedche2002measuring} evaluates ontologies based on a
“golden standard”  set. These approaches have  certain limitations when each  ontology has different
purposes, and  it is  difficult to  find a  standard ontology  as a  “golden standard”  ontology for
evaluation.

Moreover, there are many different criteria used for ontology evaluation, including Accuracy, Adaptability, Clarity, Completeness/Incompleteness, Consistency/Inconsistency, Conciseness, Computational Efficiency, Expandability, Sensitiveness, Redundancy, and Transparency \cite{vrandevcic2009ontology,staab2010handbook}. Different approaches have different groups of these criteria, such as G{\'o}mez-P{\'e}rez used a group of 5 criteria (Consistency, Completeness, Conciseness, Expandability, and Sensitiveness) \cite{gomez2004ontology}, while J. Bandeira proposed FOCA methodology with a group of 6 criteria (Completeness, Adaptability, Consistency, Conciseness, Computational Efficiency, and Clarity \cite{bandeira2016foca}.

To evaluate the proposed AgriComO ontology, this stud divides the ontology evaluation process into two parts, validation and verification tests. Ontology validation examines the developed ontology to determine whether the correct ontology has been developed. Besides, ontology verification examines the developed ontology to determine whether the ontology has been developed correctly. 

\subsubsection{Ontology Validation}

To validate the Agriculture Computing Ontology (AgriComO), two approaches are used to implement the validation process. The first approach requires answering eight \textit{Competency Questions} (CQs) (as shown in Table \ref{table:ComEvalMetrics}) \cite{bezerra2013evaluating}, while the second approach evaluates the ontology content with five criteria (Consistency, Completeness, Conciseness, Expandability, and Sensitiveness as shown in Table \ref{table:ContEvalMetrics}) \cite{gomez2004ontology}.

\begin{table}[hb!]
\selectfont\scriptsize\centering
  \renewcommand{\arraystretch}{1.2}\selectfont
  \caption{Answering Competency Questions}
  \setlength{\tabcolsep}{3pt}
  \label{table:ComEvalMetrics}%
    \begin{tabular} {|c|m{6.5cm}|c|}
    \hline
        \textbf{No.} & \textbf{Question} & \textbf{Check} \\
    \hline
        \textbf{CQ1}  	  & \textbf{What types of mined knowledge models are represented in the system?} \newline
    \textbf{Answer}: There are four types of mined knowledge models presented in the AgriComO ontology, including \textit{Classification}, \textit{Clustering}, \textit{Regression}, and \textit{Association rule}. & Passed \\
    \hline
        \textbf{CQ2} 	  & \textbf{What types of elements are represented in each mined knowledge model?} \newline
    \textbf{Answer}: Common elements of mined knowledge models in data mining, including input condition and output target attributes, algorithms, data transformation functions, dataset, evaluation metrics,  the environment of experiments, locations and several contributing factors in digital agriculture.
        & Passed \\
    \hline
        \textbf{CQ3}      & \textbf{What concepts of a given domain are represented as input/output attributes of the models?} \newline
    \textbf{Answer}: Agriculture concepts are nested under \textit{VirtualEntity} and \textit{PhysicalEntity} of the AgriComO ontology. & Passed \\
    \hline
        \textbf{CQ4} 	  & \textbf{What transformations of particular elements are used in mined knowledge models?} \newline
    \textbf{Answer}: AgriComO ontology includes data transformations and its sub-classes for domain concepts, computing algorithms its sub-classes for computing concepts. & Passed \\
    \hline
        \textbf{CQ5}      & \textbf{What types of relationships are represented in the system?} \newline
    \textbf{Answer}: AgriComO ontology includes common relationships between agriculture concepts, between data mining concepts, and between agriculture and data mining concepts. & Passed \\
    \hline
        \textbf{CQ6}      & \textbf{How mined knowledge models represent the data values of model elements?} \newline
    \textbf{Answer}: AgriComO ontology includes class \textit{State} to represent state values. & Passed \\
    \hline
        \textbf{CQ7}      & \textbf{How mined knowledge models record, transform, and transmit data?} \newline
    \textbf{Answer}: Based on AgriComO ontology, mined knowledge models are captured as a set of concepts, relations, transformations and values, then transformed into the OAK KMaps  Repository. & Passed \\
    \hline
        \textbf{CQ8}      & \textbf{What types of mined knowledge models arise from the external aspects of the system and its improvement stages?} \newline
    \textbf{Answer}: The type of mined knowledge models can be extended with other data mining tasks, such as data/attribute selection.
        & Passed \\
    \hline
  \end{tabular}%
\end{table}%

In the first test of ontology validation, AgriComO ontology is evaluated based on eight CQs for a competency evaluation. CQs not only specify the requirements for the ontology in the ontology development lifecycle but also provide references for a competency evaluation. 
In the evaluation step, answering eight CQs in the list of CQs supports reviewing the proposed ontology with its requirements. It ensures that the final ontology has enough competency for use in the model. Each question is reviewed to provide the answer and make the decision \textit{Passed} or \textit{Failed}. The result for each CQ is \textit{Passed} if the answer matches its assumption. If it does not match, the result is \textit{Failed}. The ontology will pass the competency evaluation of this validation test if all CQs have answers and are graded as \textit{Passed}.

Basically, eight CQs of the list have contributed to the ontology development progress and each step also aims to meet particular CQs. So, all CQs have results \textit{Passed}. Table \ref{table:ComEvalMetrics} provides a summary of answers for the competency evaluation.

\newcommand{\ssymbol}[1]{^{\@fnsymbol{#1}}}
\makeatother

\begin{table}[htb]
\selectfont\scriptsize\centering
  \renewcommand{\arraystretch}{1}
  \caption{Content Evaluation Metrics}
  \setlength{\tabcolsep}{2.5pt}
  \label{table:ContEvalMetrics}%
    \begin{tabular} {|c|c|m{4.9cm}|c|}
    \hline
        \textbf{No.} & \textbf{Criteria}        & \textbf{Explanation}   & \textbf{Result} \\
    \hline
        \textbf{C1}  & \textbf{Consistency}     & AgriComO ontology contains all over 600 concepts and 1,300 transformations, which are consistent.   	                              & Yes\\
        \textbf{C2}  & \textbf{Completeness} 	& AgriComO ontology reflects an image for scope of OAK model for representing data mining results in digital agriculture.   	          & Yes\\
        \textbf{C3}  & \textbf{Conciseness}     & AgriComO ontology is free of any needless concepts or redundancies between concepts. & Yes\\
        \textbf{C4}  & \textbf{Expandability} 	& AgriComO ontology is a well-defined and scalable ontology.   	                                                  & Yes\\
        \textbf{C5}  & \textbf{Sensitiveness}   & Small changes in AgriComO ontology are not observant for the current concepts.   	  & Yes\\
    \hline
  \end{tabular}%
\end{table}%

In the second test, the content of ontology is evaluated based on 5 criteria, including Consistency, Completeness, Conciseness, Expandability, and Sensitiveness (proposed by G{\'o}mez-P{\'e}rez \cite{gomez2004ontology}). Each criterion is evaluated manually to answer \textit{Yes} or \textit{No}. The ontology will pass the content evaluation of this validation test if all criteria have answered \textit{Yes}.
After checking AgriComO ontology with all five criteria as well as their guides, the proposed ontology has the answer \textit{Yes} for all criteria as shown in Table \ref{table:ContEvalMetrics}.

After the first part of the ontology evaluation, the AgriComO ontology has been proved the ability to be used in the OAK model as a core background knowledge for OAK KMaps Repository. It is also evaluated with 5 criteria to ensure that the ontology development is carried on correctly.

\subsubsection{Ontology Verification}
\label{sec:OntVerify}

There are two approaches for ontology verification, including ontology taxonomy evaluation and FOCA methodology evaluation. 
In the first verification test, AgriComO ontology is evaluated based on the taxonomy of ontology. For ontology taxonomy evaluation, Lovrencic and Cubrilo proposed a set of criteria for evaluating the taxonomy \cite{lovrencic2008ontology}. The taxonomy evaluation approach has three criteria (including Inconsistency, Incompleteness, and Redundancy). These three criteria are divided into seven sub-criteria (including circularity errors, definition errors, semantic errors, incomplete concept classification, partition errors, grammatical redundancy, and identical formal definition). These criteria and sub-criteria intend to find existing errors in the ontology. Each sub-criteria is verified to answer \textit{Yes} or \textit{No}. The ontology will pass the taxonomy evaluation of this verification test if all sub-criteria have answered \textit{No}.
For this test, AgriComO ontology has been reviewed to find inconsistency, incompleteness and redundancy errors based on the seven sub-criteria above. The summary result of this evaluation is presented in Table \ref{table:TaxoEvalMetrics}.

\begin{table}[!htb]
\selectfont\scriptsize\centering
    \renewcommand{\arraystretch}{1.2}
  \caption{Taxonomy Evaluation Metrics}
  \label{table:TaxoEvalMetrics}%
    \setlength{\tabcolsep}{2.5pt}
    \begin{tabular}{|m{3cm}|m{4cm}|c|}
    \hline
        \textbf{Criteria}  & \textbf{Explanation}  & \textbf{Check} \\
    \hline
        \multicolumn{3}{|l|}{\textbf{C1. Inconsistency}} \\
    \hline
        SC1. Circularity errors & All concepts are stated as specialisation themselves. & No\\
    \hline
        SC2. Definition errors & No wrongly defines concepts.           & No \\
    \hline
        \multicolumn{3}{|l|}{\textbf{C2. Incompleteness}} \\
    \hline
        SC3. Semantic errors  & Incorrect semantic classification.      & No \\
    \hline
        SC4. Incomplete concept classification & All concepts are overlooked by classification. & No \\
    \hline
        SC5. Partition errors & A partition between a set of concepts is omitted. & No\\
    \hline
        \multicolumn{3}{|l|}{\textbf{C3. Redundancy}} \\
    \hline
        SC6. Grammatical redundancy  & More than one explicit definition.   & No \\
    \hline
        SC7. Identical formal definition  & Concepts with same formal definition. & No \\
    \hline
  \end{tabular}%
\end{table}%

The result in Table \ref{table:TaxoEvalMetrics} has shown that there is no inconsistency, incompleteness or redundancy error in the AgriComO ontology. This result reflects the detail and carefulness of the ontology development process. Every concept from the digital agriculture domain and computing domain has been reviewed and located in the ontology.

In the second verification test, the AgriComO ontology is evaluated based on the FOCA approach with the GQM metric \cite{bandeira2016foca}. This test includes five goals (including Substitude, Ontological Commitments, Intelligent Reasoning, Efficient Computation and Human Expression) and six criteria (aka Metric in FOCA methodology) (including Adaptability, Completeness, Consistency, Computational Efficiency, Conciseness, and Clarity). To evaluate them, a set of 13 questions is used to review the ontology and grade. This is a quantitative evaluation process, the questions are evaluated with a maximum score of 100. The total quality score is calculated from the score of 4 goals (Substitude, Ontological Commitments, Intelligent Reasoning, and Efficient Computation) and the experience of the evaluator as follows:

\footnotesize{
\begin{equation*}
\label{eq:FOCA}
\begin{aligned}
    \mu _i = 
    \begin{matrix} 
        \frac{
                \begin{matrix}
                    exp\{-0.44 + 0.03(Cov_s\times Sb)_i + 0.02(Cov_C\times Co)_i \\ 
                     + 0.01(Cov_R \times Re)_i + 0.02(Cov_{CP} \times Cp)_i \\
                     - 0.66LExp_i - 25(0.1 \times Nl)_i\}
                \end{matrix}
            }
            {
                \begin{matrix}
                    1 + exp\{-0.44 + 0.03(Cov_s\times Sb)_i + 0.02(Cov_C\times Co)_i  \\
                    + 0.01(Cov_R \times Re)_i+ 0.02(Cov_{CP} \times Cp)_i \\
                    - 0.66LExp_i - 25(0.1 \times Nl)_i\}
                \end{matrix}
            }
    \end{matrix}
\end{aligned}
\end{equation*}
}
\normalsize
where, \\
  * \(Cov_s\) is the mean of grades obtained from G1 (Substitude)\\
  * \(Cov_c\) is the mean of grades obtained from G2 (Ontological Commitments)\\
  * \(Cov_R\) is the mean of grades obtained from G3 (Intelligent Reasoning)\\
  * \(Cov_{Cp}\) is the mean of grades obtained from G4 (Efficient Computation)\\
  * \(LExp\) corresponds with the experience of the evaluator (1 for experienced evaluator in ontologies, 0 if otherwise)\\
  * \(Nl\) is 1 only if some Goal was impossible for the evaluator to answer all the questions, 0 if otherwise\\
  * \(Sb,\ Co,\ Re,\ Cp\) is 1 if total quality score considers Goal G1, G2, G3, G4 respectively, 0 if otherwise.
In this approach, the total quality score \(\mu_i\) does not contain the calculated mean of the goal G5, however, the mean of the G5 shows the clarity criteria of the ontology, which helps scientists during identifying concepts in the process.

\begin{table}[htb!]
\selectfont\scriptsize\centering
  \renewcommand{\arraystretch}{1.2}
  \caption{Verification of Questions in the GQM Model}
  \setlength{\tabcolsep}{3pt}
  \label{table:VerifyQs}%
    \begin{tabular}{|c|c|m{5.4cm}|c|}
    \hline
        \textbf{Goal} & \textbf{Question} & \textbf{Explanation} & \textbf{Grade} \\
    \hline
        \textbf{G1} & \textbf{Q1} & The document defines the ontology objective, the ontology stakeholders, and the use of scenarios & 100 \\
        \textbf{G1} & \textbf{Q2} & Eight CQs are provided and answered clearly. & 100 \\
        \textbf{G1} & \textbf{Q3} & The ontology reused several concepts from the SSN Ontology, Plant Ontology, etc. & 25 \\
    \hline
        \textbf{G2} & \textbf{Q4} & The type of the ontology is application ontology. The ontology has moderate abstraction concepts.  & 50 \\
        \textbf{G2} & \textbf{Q5} & Did the ontology impose a maximum ontological commitment?  & - \\
        \textbf{G2} & \textbf{Q6} & The ontology properties reflect characteristics of concepts in the agriculture domain.  & 100 \\
    \hline
        \textbf{G3} & \textbf{Q7} & One of mechanisms for building the ontology is top-down mechanism, therefore, there are no contradictions. & 100 \\
        \textbf{G3} & \textbf{Q8} & There are no redundancies.  & 100 \\
    \hline
        \textbf{G4} & \textbf{Q9} & The reasoner Pellet 1.5.2 (Protege plugin) has been run and check the consistency of the ontology (classes and instances with their properties, characteristics, and constraints).  & 100 \\
        \textbf{G4} & \textbf{Q10} & The ontology can be stored as OWL, RDF format, and imported into RDF SPARQL Endpoint for query and reasoning.  & 100 \\
    \hline
        \textbf{G5} & \textbf{Q11} & The documentation provides modelling samples of data mining tasks for agriculture. & 100 \\
        \textbf{G5} & \textbf{Q12} & Concepts (classes and instances) are described by \textit{dc:description} relations. & 100 \\
        \textbf{G5} & \textbf{Q13} & All concepts are provided at lease the \textit{dc:identifier}, \textit{rdfs:lable}, and \textit{dc:description} relations to define their definitions.  & 100 \\
    \hline
  \end{tabular}%
\end{table}%

According to the FOCA methodology, there are two types of ontology quality verification, partial quality verification and total quality verification. For the AgriComO ontology, the total quality verification method was chosen because the quality of the ontology in this method considers all five roles of knowledge representation. After reviewing and grading each question in the list of 13 questions, the summary result shows in Table \ref{table:VerifyQs}. Applying the above equation, the total quality verification for AgriComO ontology is calculated as follows:

\footnotesize{
\begin{equation*}
\label{eq:FOCAscore}
\begin{aligned}
    \mu _i & =
    \begin{matrix} 
    \frac{
            \begin{matrix}
                exp\{-0.44 + 0.03(Cov_s\times Sb)_i + 0.02(Cov_C\times Co)_i \\ 
                \;\;\;\;+ 0.01(Cov_R \times Re)_i + 0.02(Cov_{CP} \times C_p)_i \\
                - 0.66LExp_i - 25(0.1 \times Nl)_i\}
            \end{matrix}
        }
        {
            \begin{matrix}
                1 + exp\{-0.44 + 0.03(Cov_s\times Sb)_i + 0.02(Cov_C\times Co)_i  \\
                \;\;\;\;\;\;\;\;\;\;\,+ 0.01(Cov_R \times Re)_i + 0.02(Cov_{CP} \times Cp)_i \\
                -0.66LExp_i-25(0.1 \times Nl)_i\}
            \end{matrix}
        }
    \end{matrix} \\
    & =
    \begin{matrix} 
    \frac{
            \begin{matrix}
                exp\{-0.44 + 0.03(Cov_s)_i + 0.02(Cov_C)_i \\ 
                + 0.01(Cov_R)_i + 0.02(Cov_{CP})_i - 0.66LExp_i\}
            \end{matrix}
        }
        {
            \begin{matrix}
                1 + exp\{-0.44 + 0.03(Cov_s)_i + 0.02(Cov_C)_i \\
                + 0.01(Cov_R)_i + 0.02(Cov_{CP})_i - 0.66LExp_i\}
            \end{matrix}
        }
    \end{matrix} \\
    & = 0.95
\end{aligned}
\end{equation*}}
\normalsize
where, \\
* \(Nl=0\) because the questions of all goals were answered. \\
* \(Sb=1,\ Co=1,\ Re=1,\ Cp=1\) because the total quality considers all the roles.\\
The result of the total quality of the proposed ontology is 0.95 and it is very close to 1, which means that this ontology has high quality and satisfies the requirements of an agriculture computing ontology used in the proposed OAK model. Moreover, the ontology also obtained six criteria from the FOCA evaluation, including Completeness, Adaptability, Conciseness, Consistency, Computational efficiency, and Clarity.

Overall, the AgriComO ontology has passed four ontology evaluation tests, including validation evaluation and verification evaluation. Passing four evaluation tests, the AgriComO has been verified with total 12 criteria, including Adaptability, Consistency, Completeness, Conciseness, Computational efficiency, Clarity, Expandability, Sensitiveness, Inconsistency, Incompleteness, and Redundancy.
\subsection{Knowledge Browser Performance}
\label{sec:PracticalEVAL}

To demonstrate the ability of the Knowledge Repository and Knowledge Browser, this study designs practical experiments in different end-user groups: data scientists and agronomic users.

Although agronomists and data scientists have a wide range of concerns in digital farming, these concerns can be classified into several tasks of data mining in digital agriculture, such as yield prediction, early disease prediction, or nitrogen content estimation. Agronomists and agricultural production managers from Origin Enterprises PLC\footnote{Origin Enterprises Public Limited Company is a company focused on consumer foods, crop nutrition, feed ingredients, marine proteins, and oils. This company has manufacturing and distribution operations in Ireland, the United Kingdom and Poland. Origin Enterprises PLC is also an enterprises partner of CONSUS project.} also provided a list of the greatest potential to gain from precision farming and digital agriculture as follows \cite{agrii2020farm}:
\begin{itemize}
      \item Saving costs based on limited inputs
      \item Finding crop yield potential earlier
      \item Identifying farming problems earlier
      \item Providing a map of production costs and gross margin
      \item Improving performance from every part of the field
      \item Developing tools for support decision making
      \item Reducing human errors from farming progress
      \item Reducing time consumption on farming activities
\end{itemize}
In the list above, the last 2 bullet points are heavily related to agriculture machinery, while the rest of the list are similar problems of data mining in digital agriculture with different factors of interest. Based on these concerns, this study has created a list of common queries from data scientists working on digital agriculture and agronomists using data mining results. Again, this list contains a lot of questions, but they are compiled into a list of 10 queries, which can be delegated to a wide number of similar questions. They are as follows:

\textbf{Queries from data scientist users:}
\begin{itemize}
    \item \textbf{Q1.} What is the basic information about \textit{wheat crop}?
    \item \textbf{Q2.} What concepts are used in knowledge item \textit{Regressor\_0015}?
    \item \textbf{Q3.} What models can use \textit{nitrogen} to predict and what to predict?
    \item \textbf{Q4.} What models can be used to predict \textit{wheat yield}?
    \item \textbf{Q5.} What potential methods can be used to process \textit{Temperature} in knowledge items?
\end{itemize}

\textbf{Queries from agronomic users:}
\begin{itemize}
    \item (\textbf{Q1.} What is the basic information of \textit{wheat crop}?)
    \item  \textbf{Q6.} What are the relationships between \textit{Wheat} and \textit{Leaf Rust} disease?
    \item  \textbf{Q7.} What potential characteristics or states can be used to predict \textit{high yield}?
    \item  \textbf{Q8.} How crops can get a \textit{high yield} when grown in the \textit{UK}? 
\end{itemize}

\textbf{Extra queries from testing:}
\begin{itemize}
    \item \textbf{Q9.} What is relevant information of \textit{Multi-Linear Regression}?
    \item \textbf{Q10.} What are knowledge items related to dataset \textit{PlantVillage}?
\end{itemize}

Basically, using a list of 10 queries to test the knowledge management system is a practical approach for evaluating the system. It partly presents the ability of the proposed OAK model in handling mined knowledge in digital agriculture. However, this section will analyse and prove that the list of these 10 queries is enough to access all elements and roles of the OAK model. Other different queries can be inherited and generated from these ten queries.

In general, several questions can be answered by the first layer of knowledge, the AgriComO ontology. For example, queries Q1, Q5, Q6, and Q9 can be partly answered by the knowledge in AgriComO ontology. However, all the above queries can be completely answered by the OAK model when it has enough knowledge in the knowledge repository. Moreover, only KMaps can help to answer questions Q4, Q7, and Q8 effectively.

According to Definition \ref{def:KM} Knowledge Map Model of OAK Model (Section \ref{sec:OAK}), there are 5 elements that contributed to knowledge representations, including Concept, Instance, Relation, Transformation, and State (Remark: \(\mathbb{KM} = (\mathbb{C}, \mathbb{I}, \mathbb{R}, \mathbb{T}, \mathbb{S})\)). Definition \ref{def:KE} Extend Knowledge of OAK Model (Section \ref{sec:def4KRep}) also provides 8 types of instances presented in knowledge representations, including KMap, Algorithm, Condition, Target, Dataset, Evaluation, Location, and Context). Therefore, these experiments will evaluate the accuracy in the retrieval of different elements in different roles of instances based on given queries as follows:
\begin{itemize}
    \item \textbf{Elements in OAK model}: Concept, Instance, Relation, Transformation, and State (Section \ref{sec:OAK}).
    \item \textbf{Roles of Instances}: KMap, Algorithm, Condition, Target, Dataset, Evaluation, Location, and Context (Section \ref{sec:def4KRep}).
\end{itemize}

To demonstrate the effectiveness of the proposed Knowledge Repository, this study carries out a set of 10 queries, which will focus on looking up basic knowledge or expert knowledge (Q1, Q5, Q6, and Q9) and finding mined knowledge items from existing relevant data mining studies (Q2, Q3, Q4, Q7, Q8, and Q10). These experiments analyse queries to provide the solution and SPARQL queries, which can execute in SPARQL Endpoint to have results. All queries are formalised, generated SPARQL queries and then executed in SPARQL Endpoint. They are also searched on Knowledge Browser with corresponding keywords. 
For example, query Q3 is processed as follows:

\textbf{Query}: \textbf{Q3.} What models can use nitrogen to predict and what to predict? \\
\textbf{Formalise}: \textbf{QF3.} What are knowledge items \(M_i\), which use condition \(c\) (for example, \textit{Nitrogen}), to predict? [\textit{DM models}] \\
\textbf{Solution}: Finding all instances, which have relation \textit{hasCondition} and objects, which are instances of concept \textit{Nitrogen}, then query all information related to returned results (Listing \ref{lstQ3}).
    
\begin{lstlisting}[caption=SPARQL Query for QF3, label=lstQ3]
PREFIX rdf: <http://www.w3.org/1999/02/22-rdf-syntax-ns#> 
PREFIX AgriComO:  <http://www.ucd.ie/consus/AgriComO#> 
PREFIX AgriKMaps:   <http://www.ucd.ie/consus/AgriKMaps#> 
SELECT  ?subject
WHERE {
    ?subject  AgriComO:hasCondition ?object .
    ?object   rdf:type          AgriComO:Nitrogen .
}
\end{lstlisting}

\begin{figure}[htb!]
    \centering
    \includegraphics[width=8.5cm]{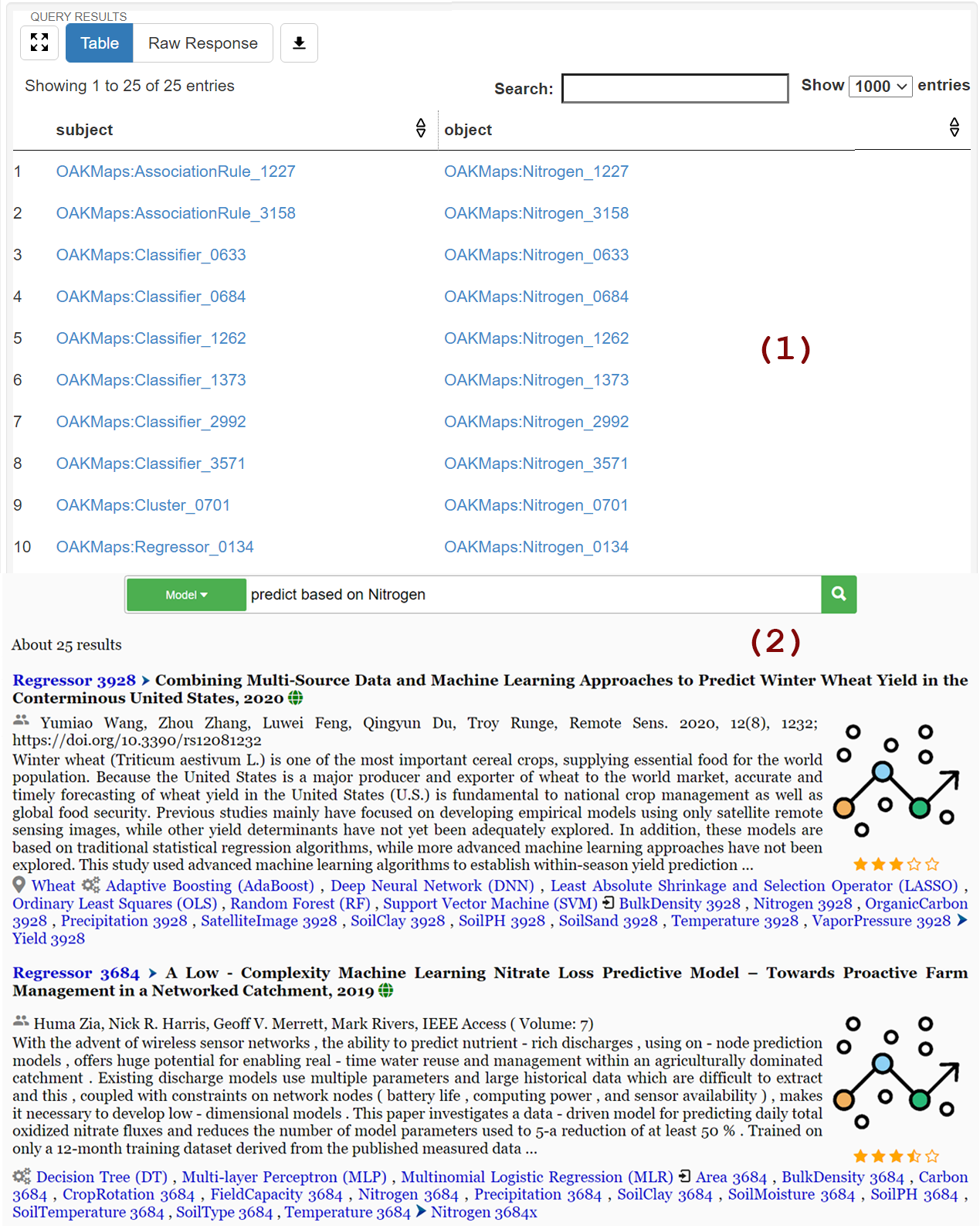}
    \caption{Results of QF3 on SPARQL Endpoint (1) and on Knowledge Browser (2).}
    \label{fig:Q3}
\end{figure}

Listing \ref{lstQ3} shows the SPARQL query for QF3. It is executed on SPARQL Endpoint and got 25 knowledge items from AgriKMaps Respository (Figure \ref{fig:Q3}, (1)). However, this SPARQL query can expand to have more information on related elements. They are processed to present as a web page in Knowledge Browser with query "predict based on Nitrogen" (Figure \ref{fig:Q3} (2)).

To evaluate the correctness and completeness of the proposed OAK model, we created an access matrix to review the representation and access ability of the model (as shown in Table \ref{table:CrossCheckQUERY}). This table includes five different elements of OAK Model in eight different roles. Moreover, these elements and roles are considered in two different groups when executing queries, including input parameters and output results. For each query (from Q1 to Q10 for two groups of experiments), each element in the OAK model and each role of instances will be checked (\checkmark) if it is used, accessed, or returned in the results. Column \textit{Check} at the last column shows the result of each item based on the whole list of queries. It is checked (\checkmark) if it has at least one check from any query.

\begin{table}[htb!]
\selectfont\scriptsize\centering
  \renewcommand{\arraystretch}{1.1}
  \caption{Access Matrix based on Different Elements and Roles for Each Query}
  \label{table:CrossCheckQUERY}%
    \setlength{\tabcolsep}{2.5pt}
    \begin{tabular}{lccccccccccc}
    \hline
        \textbf{Element} & \textbf{Q1} & \textbf{Q2} & \textbf{Q3} & \textbf{Q4} & \textbf{Q5} & \textbf{Q6} & \textbf{Q7} & \textbf{Q8}  & \textbf{Q9}  & \textbf{Q10} & \textbf{Check}\\
    \hline
        \multicolumn{11}{l}{\textbf{Input Element}} \\
        Concept	        &\checkmark&	 &\checkmark&\checkmark&\checkmark&  &\checkmark& & \checkmark  & \checkmark & \checkmark \\
        Instance	    &  &\checkmark&	 &	 &	  &	 & & & \checkmark  & \checkmark & \checkmark \\
        State	        &  &  &  &\checkmark&	 &  &\checkmark&\checkmark&  & & \checkmark\\
        Transformation&  &  &  &  &	 &\checkmark&	  & & \checkmark & \checkmark & \checkmark \\
        Relation	    &  &  &	\checkmark &\checkmark&\checkmark&\checkmark& \checkmark& & \checkmark  & \checkmark & \checkmark \\
    \hline\noalign{\smallskip}
        \multicolumn{11}{l}{\textbf{Input Role}}\\
        KMap	        &  &\checkmark& & & & &	 & & & & \checkmark \\
        Algorithm	    &  &  &	 &  & &	& & & \checkmark & \checkmark & \checkmark\\
        Condition	    &  &  &	\checkmark &  &	 & &	 &   & & & \checkmark\\
        Target        &  &  &  &\checkmark&	 & &\checkmark&\checkmark& & & \checkmark \\
        Dataset	    &  &  &	 &  &	& &	& & \checkmark  & \checkmark & \checkmark\\
        Evaluation	&  &  &	 &  &	& & & & \checkmark & \checkmark& \checkmark\\
        Location	    &  &  &	 &  &	& & &\checkmark& \checkmark & \checkmark & \checkmark\\
        Context	    &  &  &	 &&	 &	& \checkmark &\checkmark& & & \checkmark\\
    \hline
        \multicolumn{11}{l}{\textbf{Output Element}}\\
        Concept	        & \checkmark & \checkmark & \checkmark &\checkmark &\checkmark 
                        &  \checkmark & \checkmark &\checkmark & \checkmark &\checkmark 
                        & \checkmark \\
        Instance	    & \checkmark & \checkmark & \checkmark & \checkmark & 
                        & & \checkmark & \checkmark & \checkmark & \checkmark 
                        & \checkmark \\
        State	        & \checkmark & \checkmark & \checkmark & \checkmark & 
                        & & \checkmark & \checkmark & \checkmark & \checkmark 
                        & \checkmark\\
        Transformation& \checkmark & \checkmark & \checkmark & \checkmark & \checkmark 
                        & \checkmark & \checkmark & \checkmark & \checkmark & \checkmark 
                        &\checkmark \\
        Relation	    & \checkmark & \checkmark & \checkmark & \checkmark & \checkmark 
                        & \checkmark & \checkmark & \checkmark & \checkmark & \checkmark 
                        & \checkmark \\
    \hline\noalign{\smallskip}
        \multicolumn{11}{l}{\textbf{Output Role}} \\
        KMap	        &  &\checkmark & \checkmark & \checkmark &	& &\checkmark&\checkmark& & \checkmark & \checkmark \\
        Algorithm	    &  &\checkmark & \checkmark&\checkmark &	& &\checkmark  &\checkmark & & \checkmark & \checkmark\\
        Condition	    &  &\checkmark & \checkmark&\checkmark &	& &\checkmark  &\checkmark & & \checkmark & \checkmark\\
        Target        &  &\checkmark & \checkmark&\checkmark &	& &\checkmark  &\checkmark & & \checkmark & \checkmark\\
        Dataset	    &  &\checkmark & \checkmark&\checkmark &	& &\checkmark  &\checkmark & & \checkmark & \checkmark\\
        Evaluation	&  &\checkmark & \checkmark&\checkmark &	& &\checkmark  &\checkmark & & \checkmark & \checkmark\\
        Location	    &  &\checkmark & \checkmark&\checkmark &	& &\checkmark  &\checkmark & & \checkmark & \checkmark\\
        Context	    &  &\checkmark & \checkmark&\checkmark &	& &\checkmark  &\checkmark & &\checkmark & \checkmark\\
    \hline
        \multicolumn{12}{l}{\checkmark If the element or role contributes into the SPARQL query.}
    \end{tabular}
\end{table}%

The result of this reviewing process has shown in Table \ref{table:CrossCheckQUERY}. It can be easily seen that every element of the OAK model and every role of instances have been used, accessed, or returned in the results based on the set of ten queries (from Q1 to Q10 above). Moreover, queries for knowledge items easily access and get results of all elements of the OAK model and instances (as attributes of the prediction knowledge models) with different roles.

In the list of queries, there are two typical queries that results contain all the elements of the OAK Model. Specifically, query Q1 shows information of all related elements (whole 5 elements in the OAK model) of the input concept. This query requests domain information from the AgriComO ontology, therefore, its result do not contain knowledge items (as shown in column Q1 of Table \ref{table:CrossCheckQUERY}). The second query is query Q2, which returns information of all related elements and roles (whole 5 elements and 8 roles of instances (if exist) in the OAK model) of the input knowledge item (as shown in column Q2 of Table \ref{table:CrossCheckQUERY}). Search queries returned a list of matched knowledge instances. From this list, all related elements/roles can be accessed when expending the SPARQL queries, such as Query Q3, Q4, Q7, Q8, and Q10. Although the list of queries has 10 queries, they have accessed all elements as well as roles of instances presented in the proposed OAK model. They also retrieved different types of knowledge successfully in the Knowledge Repository.

In this study, there are over 500 mined knowledge items, which are prepared in Section \ref{sec:KnowMaterials}. All of them can be accessed by using Query 2 (QF2) and show all of their information. All 500 mined knowledge items are located in class \textit{AgriComO:DataMining} and its four sub-classes (corresponding to 4 data mining tasks) in the AgriKMaps Repository.
The results of search queries are based on this set of knowledge items. All of them can be accessed by using QF2 and show all of their information. As defined in Definition \ref{def:KE} Extend Knowledge in Section \ref{sec:def4KRep}, knowledge items will have \textit{Algorithm}, \textit{Condition}, \textit{Target}, \textit{Dataset}, \textit{Context}, \textit{Location}, \textit{Dataset}, and \textit{Evaluation}. Queries for knowledge items can expand with more detailed requirements. For example, Query Q4 will extend with \textit{Context} of the knowledge items.

The  results in  Table  \ref{table:CrossCheckQUERY}  show that  the  Knowledge  Repository with  the
processing ability of SPARQL  supports accessing all different elements of the OAK  model as well as
all different roles in  knowledge representations. Moreover, it also illuminates  that the OAK model
accurately processes, stores, and represents the mined knowledge items.

The results of queries demonstrate the performance of the OAK model and the knowledge repository and
provide ideas  for further  works. For  example, query  Q3 or query  Q10 create  a group  of similar
knowledge items.

\section{Conclusion}
\label{sec:Conclusion}

While  significant research  work has  been committed to knowledge  extraction from  large datasets,
little effort has been  devoted to knowledge reuse and mining. We presented  a generalised model for
knowledge representation, storage, and exploration.  The model, called ontology-based knowledge map,
has been demonstrated in  the agricultural domain, but it can be extended  to many other application
domains. The model  consists of eight primary  and two secondary components.  The  model is flexible
enough to represent any knowledge and mined results. We also presented a formal model justifying the
components and the process of building such a model.

Furthermore, we  presented a complete  architecture for the  ontology-based knowledge map  model and
gave numerous examples about how to represent mined knowledge in digital agriculture. In addition to
the  core  modules,  the  architecture  includes knowledge  miners,  knowledge  wrappers,  knowledge
publishing, and knowledge browser modules based on a pre-defined ontology.

We have implemented  a fully operating prototype  around an agricultural ontology  to provide domain
knowledge and  a knowledge management  system.  The system  stores knowledge and  supports efficient
knowledge exploration and retrieval. The current implementation is very adaptive and easy to use.

The prototyped ontology-based knowledge map model is  promising and has the potential to be extended
to  other application  domains. Despite  its flexibility,  efficient storage  system, and  knowledge
retrieval, the proposed model relies heavily on the quality of the inputs and repositories (previous
ontology). We plan to  design an evaluation methodology to evaluate not  only the model performance,
its robustness  and scalability,  but also  the quality of  input knowledge  items of  the knowledge
repository.

\paragraph{\textbf{Acknowledgment}} This work is part of CONSUS and is supported by the SFI Strategic Partnerships Programme (16/SPP/3296) and is co-funded by Origin Enterprises Plc.

\bibliographystyle{cas-model2-names}

\bibliography{ObasedKMap}

\bio{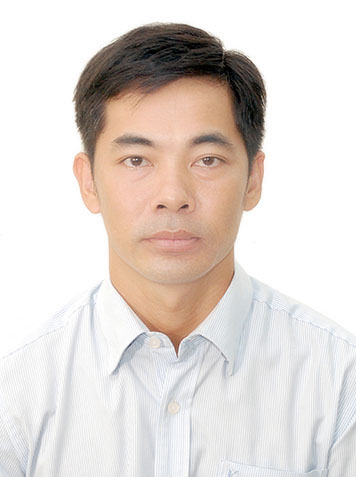}
	Quoc Hung Ngo was awarded PhD degree in Computer Science from University College Dublin (UCD). He received a Master degree in Computer Science from University of Science, Vietnam National University - Ho Chi Minh City (VNUHCM), Vietnam. He has involved in the BioCaster project by building geographical ontology, integrating the geo-ontology into the Global Health Monitor system, and building the webpage for publishing project results. His research interests are knowledge management, data analytics, semantic web and natural language processing. Currently, he is a lecturer in Computing at Technological University Dublin (TUDublin).
\endbio

\bio{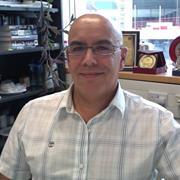}
	Tahar Kechadi was awarded PhD and Masters degree in Computer Science from University of Lille 1, France. He joined the UCD School of Computer Science (CS) in 1999. He is currently a Professor of Computer Science at CS, UCD. His research interests span the areas of Data Mining, distributed data mining heterogeneous distributed systems, Grid and Cloud Computing, and digital forensics and cyber-crime investigations. He is currently an editorial board member of the Journal of Future Generation of Computer Systems. He is a member of the communication of the ACM journal and IEEE computer society.
\endbio

\bio{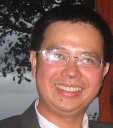}
	Nhien-An Le-Khac received the Ph.D. degree in computer science from the Institut National Polytechnique de Grenoble (INPG), France, in 2006. He was a Research Fellow with Citibank, Ireland (Citi). He is currently a lecturer with the School of Computer Science (CS), University College Dublin (UCD), Ireland. He is also the Programme Director of the UCD M.Sc. Programme in forensic computing and cybercrime investigation and an International Programme for law enforcement officers specialising in cybercrime investigations. He is also the Co-Founder of the UCD-GNECB Postgraduate Certificate in fraud and e-crime investigation.
\endbio

\clearpage
\flushend

\end{document}